%% file: acl_latex.tex
\pdfoutput=1
\documentclass[11pt]{article}

\usepackage[final]{acl}

\usepackage{times}
\usepackage{latexsym}
\usepackage{url}

\usepackage[T1]{fontenc}
\usepackage[utf8]{inputenc}

\usepackage{microtype}
\usepackage{inconsolata}

\usepackage{tabularx}
\usepackage{times}
\usepackage{latexsym}

\usepackage{graphicx}
\usepackage{graphics}
\usepackage{booktabs}
\usepackage{enumitem}
\usepackage{multirow}
\usepackage{multicol}
\usepackage{tabularx}

\usepackage{svg}
\usepackage{float}
\usepackage{hyperref}
\usepackage{amsmath}
\usepackage{amssymb} 
\usepackage{booktabs}
\usepackage{float}
\usepackage{xspace}
\usepackage{booktabs}
\usepackage{multirow}
\usepackage{multicol}

\usepackage{subcaption}
\usepackage{times}
\usepackage{tabularx}
\usepackage{longtable}
\usepackage{array}
\usepackage{booktabs}
\usepackage{xtab}
\usepackage{threeparttablex}
\usepackage{latexsym}
\usepackage{nicematrix,tikz}

\usepackage{epigraph}
\usepackage{microtype}
\usepackage{booktabs}
\usepackage{pifont}
\usepackage{xspace}
\usepackage{amssymb}
\usepackage{tabularx}
\usepackage{multirow}
\usepackage{subcaption}
\usepackage{graphicx}

\usepackage{adjustbox}
\usepackage{booktabs}
\usepackage{babel}
\usepackage{ccicons}

\usepackage{placeins}  
\usepackage{amsthm}

\theoremstyle{definition}
\newtheorem{definition}{Definition}[section]

\usepackage{xcolor}
\newcommand\dataset{\texttt{NLPContributions}}
\newcommand\datasetauto{\texttt{NLPContributions-Auto}}
\title{The Nature of NLP: Analyzing Contributions in NLP Papers}

\author{
Aniket Pramanick\textsuperscript{1}, Yufang Hou\textsuperscript{2,3}, Saif M. Mohammad\textsuperscript{4}, Iryna Gurevych\textsuperscript{1} \\
\textsuperscript{1}Ubiquitous Knowledge Processing Lab (UKP Lab) \\
Department of Computer Science and Hessian Center for AI (hessian.AI) \protect\\
Technische Universität Darmstadt \\
\textsuperscript{2}IT:U Interdisciplinary Transformation University Austria \protect\\
\textsuperscript{3}IBM Research Europe - Ireland \protect\\
\textsuperscript{4}National Research Council Canada \\
\small{\url{www.ukp.tu-darmstadt.de}, \url{yufang.hou@it-u.at}, \url{saif.mohammad@nrc-cnrc.gc.ca}}
}

\begin{document}
\maketitle

\input{sections/abstract}

\input{sections/01_intro}

\input{sections/02_rw}
\input{sections/03_dataset}
\input{sections/04_exp}

\input{sections/05_analysis}

\input{sections/06_conc}

\input{sections/07_future}

\input{sections/limit}

\input{sections/ethics}

\input{sections/ack}

\bibliography{custom}

\appendix

\input{sections/appendix}

\end{document}

%% file: sections/abstract.tex
\begin{abstract}
Natural Language Processing (NLP) is an established and dynamic %, interdisciplinary 
field. % that integrates intellectual traditions from computer science, linguistics, %social science, 
%and more. 
Despite this, %its established presence, 
%the definition of 
what constitutes NLP research remains debated. 
In this work, we address the question by quantitatively examining NLP research papers. We propose a taxonomy of research contributions and introduce \dataset{}, a dataset of nearly $2k$ NLP research paper abstracts, carefully annotated to identify scientific contributions and classify their types according to this taxonomy. We also introduce a novel task of automatically identifying contribution statements and classifying their types from research papers.
%, for which we train a strong baseline on our dataset.
We present experimental results for this task and apply our model to $\sim$$29k$ NLP research papers to analyze their contributions, aiding in the understanding of the nature of NLP research. %Our findings 
We show that NLP research has taken a winding path --- with the focus on language and human-centric studies being prominent in the 1970s and 80s, tapering off in the 1990s and 2000s, and starting to rise again since the late 2010s. Alongside this revival, we observe a steady rise in dataset and methodological contributions since the 1990s, such that today, on average, individual NLP papers contribute in more ways than ever before.
% increasingly diverse %and interdisciplinary 
% ways. 
Our dataset and analyses offer a powerful lens for tracing research trends and offer potential for generating informed, data-driven literature surveys.~\footnote{Code and data are available at:~\url{https://github.com/UKPLab/acl25-nlp-contributions}}

\end{abstract}

%% file: sections/01_intro.tex
\section{Introduction}

% \begin{figure}
%     \centering
%     \scalebox{0.50}{
%     \includegraphics[width=0.68\textwidth]{asssets/intro_fig.pdf}
%     }
%     \caption{Example of Contributions and their Roles from \citet{oprea-etal-2022-chatbot}.}
%     \label{fig:intro_fig}
% \end{figure}

% \epigraph{``He who does not know to which port he is sailing, no wind is favorable.''}{- Seneca}

Categorizing research by scientific discipline has several benefits, including bringing together scientists to make progress in a cohesive area of interest. While there is often some broad description of what a scientific discipline constitutes, the nature of a discipline is dynamic and multifaceted, and it can change with time. NLP is a particularly interesting discipline in this regard, not only because of its interdisciplinary nature, drawing on ideas and techniques from computer science, linguistics, social science, etc., but also because fundamental questions such as `\textit{what is NLP research?}' can be contentious. 
Is it the study and development of algorithms that give machines the ability to respond to and generate language? Is it the study of natural language using computational approaches? Does it cover all research at the intersection of computation and language? Or is it something more narrow?

% Regardless of how one may choose to define NLP, a separate and 
A compelling way to answer `\textit{what is NLP research?}' is to examine the papers published in NLP conferences and journals. After all, the body of current published research is the best indicator of what a field is and the nature of the field, regardless of how that field may have once been defined or understood.   

A key window into the nature of a particular research project is how the authors articulate their contributions. 
\textit{Contributions} are new scientific achievements attributed to the authors.
Roughly speaking, scientific contributions are of two types: %1.\@ 
i) those that add to human knowledge, e.g., discovering the structure of DNA; and ii)  %2.\@ 
those that create new and useful artifacts, e.g., a general-purpose chat system such as ChatGPT.
% include adding  knowledge about language or introducing new datasets \citep{sateli2015semantic}
When authors present their work in scientific papers, they describe their contributions to the research community. %it.
We define {\it contribution statements} as descriptions of these contributions. %attributed to 
% by the authors themselves.

In this paper, we propose that automatically extracting, categorizing, and quantitatively analyzing the contribution statements in the research papers of a field provides key insights into the nature of the field.
Additionally, such an effort enables historical (longitudinal) analyses of the field~\citep{sep-thomas-kuhn} and can help researchers identify emerging trends and stay current amid the rapid proliferation of scientific publications. 

We explore this idea concretely and empirically by examining $28,937$ 
NLP %research
papers published between $1974$ and $2024$. Specifically, we:

\begin{enumerate}
\itemsep-0.2em 
\item Introduce a {\it taxonomy} of contribution types common in NLP papers (\S~\ref{subsec:taxonomy}).
\item Create a {\it dataset} \dataset{} comprising of $1,995$ NLP research papers with manually annotated contribution statements and contribution types from their abstracts (\S~\ref{subsec:curation}).
% of blah papers with manually annotated contribution statements and contribution types.
\item Propose a {\it novel task} to automatically extract and classify contribution statements into contribution types from NLP papers (\S~\ref{subsec:task}).
%(a new task). .
\item Finally, ask (and answer) some preliminary questions on the {\it nature of NLP research} and how it has changed over the years (\S~\ref{sec:analysis}).
\end{enumerate}

\begin{table*}[t]
    \centering
    \scalebox{0.95}{
    \begin{adjustbox}{width=2.2\columnwidth, center}
        \begin{tabular}{l l p{9.5cm} p{14.0cm}}
        \toprule
        {\bf Type} & {\bf Sub-type} & {\bf Description} & {\bf Example}\\
        \midrule
        \multirow{5}{*}{Knowledge} & k-dataset & Describes new knowledge about datasets, such as their new properties or characteristics. & ``Furthermore, our thorough analysis demonstrates the average distance between aspect and opinion words are shortened by at least $19\%$ on the standard SemEval Restaurant14 dataset.'' --~\citet{zhou-etal-2021-closer} \\

        & k-language & Presents new knowledge about language, such as a new property or characteristic of language. & ``In modern Chinese articles or conversations, it is very popular to involve a few English words, especially in emails and Internet literature.'' --~\citet{zhao-etal-2012-part} \\

        & k-method & Describes new knowledge or analysis about NLP models or methods (which predominantly draw from Machine Learning). & ``Different generative processes identify specific failure modes of the underlying model.'' --~\citet{deng-etal-2022-model} \\

        & k-people & Presents new knowledge about people, humankind, society, or human civilization. & ``Combating the outcomes of this infodemic is not only a question of identifying false claims, but also reasoning about the decisions individuals make.'' --~\citet{pacheco-etal-2022-holistic} \\
        
        & k-task & Describes new knowledge about NLP tasks. & ``We show that these bilingual features outperform the monolingual features used in prior work for the task of classifying translation direction.'' --~\citet{eetemadi-toutanova-2014-asymmetric} \\

        % & Dataset & describes new knowledge about datasets, such as their new properties or characteristics. & ``Furthermore, our thorough analysis demonstrates the average distance between aspect and opinion words are shortened by at least $19\%$ on the standard SemEval Restaurant14 dataset.'' --~\citet{zhou-etal-2021-closer} \\
        % & Language & Presents new knowledge about language, such as a new property or characteristic of language. & ``In modern Chinese articles or conversations, it is very popular to involve a few English words, especially in emails and Internet literature.'' --~\citet{zhao-etal-2012-part} \\
        \midrule 
        \multirow{3}{*}{Artifact} & a-dataset & Introduces a new NLP dataset (i.e., textual resources such as corpora or lexicon). & ``We present a new corpus of Weibo messages annotated for both name and nominal mentions.'' --~\citet{peng-dredze-2015-named} \\

        & a-method & Introduces or proposes a new or novel NLP method or model (primarily to solve NLP task(s)). & ``The paper also describes a novel method, EXEMPLAR, which adapts ideas from SRL to less costly NLP machinery, resulting in substantial gains both in efficiency and effectiveness, over binary and n-ary relation extraction tasks.'' --~\citet{mesquita-etal-2013-effectiveness} \\
        
        & a-task & Introduces or proposes a new or novel NLP task (i.e., well-defined NLP problem). & ``We formulate a task that represents a hybrid of slot-filling information extraction and named entity recognition and annotate data from four different forums.'' --~\citet{durrett-etal-2017-identifying} \\
        
        % & Method & Introduces or proposes a new or novel NLP method or model (primarily to solve NLP task(s)). & ``The paper also describes a novel method, EXEMPLAR, which adapts ideas from SRL to less costly NLP machinery, resulting in substantial gains both in efficiency and effectiveness, over binary and n-ary relation extraction tasks.'' --~\citet{mesquita-etal-2013-effectiveness} \\
        % & Model &  Introduces new NLP model. & ``We present a new model for event extraction that jointly considers both the local context around a phrase along with the wider sentential context in a probabilistic framework.''~\citep{patwardhan-riloff-2009-unified} \\
        % & Task & Introduces or proposes a new or novel NLP task. & ``We formulate a task that represents a hybrid of slot-filling information extraction and named entity recognition and annotate data from four different forums.''~\citep{durrett-etal-2017-identifying} \\
        % & Dataset & Introduces a new NLP dataset (i.e., textual resources such as corpora or lexicon). & ``We present a new corpus of Weibo messages annotated for both name and nominal mentions.'' --~\citet{peng-dredze-2015-named} \\

        \bottomrule
        
        \end{tabular}
    \end{adjustbox}
    }
    % \caption{Description of the taxonomy on NLP research contributions with examples.}
    \caption{Overview of the taxonomy for NLP research contributions with examples for each contribution type.}
    %\caption{Annotation Scheme[TODO: improve formatting]}
    \label{tab:annot_scheme}
    % \vspace*{-4mm}
\end{table*}

%% file: sections/02_rw.tex
\section{Related Work}

\noindent{\textbf{NLP Scientometrics.}} The study of trends in scientific research gained attention following the seminal work by \citet{hall2008studying}. This line of work, broadly known as ``scientometrics'', focuses on the quantitative analysis of scholarly literature. NLP scientometrics has 
% seen significant advancements, particularly 
gained interest in recent years, as researchers strive to understand the growing landscape of NLP research and its evolution \citep{mingers2015review, chen2019visualizing}.
% NLP scientometrics has witnessed significant advancements in recent years as researchers strive to understand the landscape of NLP research and its evolution \citep{mingers2015review, chen2019visualizing}. 
One prominent research direction in NLP scientometrics is the analysis of metadata \citep{mohammad2020nlp}, employing bibliometric techniques \citep{wahle2022d3}, co-authorship analysis \citep{mohammad-2020-examining}, and topic modeling \citep{jurgens2018measuring} to gain insights into the dynamics of the field, identifying research trajectories. Text mining and deep learning techniques have also been utilized in NLP scientometrics to extract information from research papers, create structured datasets, and enable detailed analyses of the interactions among topics and their evolution \citep{prabhakaran2016predicting, tan-etal-2017-friendships, salloum2017survey, yang-li-2018-scidtb, prabhakaran2016predicting,hou-etal-2019-identification,pramanick-etal-2023-diachronic,sahinuc-etal-2024-efficient}. 
% However, some efforts have been made towards discourse relation extraction \citep{yang-li-2018-scidtb} from scientific papers, as well as studies on the evolution of topics within scientific fields \citep{prabhakaran2016predicting}, relatively little is known about the contributions. 
% While NLP scientometrics has concentrated on metadata, largely ignoring the rich contents of the paper, 
Our study delves deeper into NLP scientometrics by analyzing the content in research paper. 
%Additionally, some research on NLP entity extraction \citep{hou-etal-2021-tdmsci} addresses entities similar to our defined artifact sub-types; however, it typically does not focus on contributions nor on the field's boundaries.

%\noindent{\textbf{Citation Intent Analysis.}} 
\paragraph{Citation Intent Analysis.}
A large body of research has focused on understanding the purpose behind citations and developing classification systems for them \citep{stevens1965statistical, oppenheim1978highly, garzone2000towards, teufel-etal-2006-automatic, dong2011ensemble, jurgens-etal-2018-measuring}. While citation intents signal the purpose of a citation, such as providing background information or making comparisons, contributions differ as they present novel additions that a research paper introduces to its field. Citation intents may indirectly reflect a paper's contributions from the perspective of citing papers; our focus is on contributions as articulated by the authors themselves within their own work.

% Additionally, while much of citation analysis research has explored how papers reference or incorporate prior work, there has been limited investigation into the contributions made by research papers themselves. In this work, we look into the contributions made by the research papers themselves. 
%On the contrary, in this work, we study the contributions of research papers and examine how they are juxtaposed with the limitations of previous works. 

%\noindent{\textbf{Semantic Content Structuring.}} 
% Researchers have investigated structured semantic content modeling to improve scientific document searches. 
%\paragraph{Semantic Content Structuring.}
\paragraph{NLP Contribution Graph.}
\citet{d2020nlpcontributions} introduced an annotation scheme to identify \textit{information units} in scientific documents related to contributions, focusing on artifacts like models, datasets, or baselines linked to a pre-defined set of NLP tasks. \citet{dsouza-etal-2021-semeval} employed this annotation scheme to construct a knowledge graph that connects these artifact information units across NLP tasks. Deep learning methods have been applied to automate the extraction of this information units~\citep{gupta2021contrisci, gupta2024bert}. It is important to note that these units are not necessarily novel contributions from the papers they are extracted from. Unlike these efforts, our work broadens the scope by identifying and categorizing contribution statements from research papers without limiting them to specific NLP tasks. Additionally, our approach encompasses contributions that expand knowledge as well as introduce new artifacts.
% \citet{d2020nlpcontributions} proposed an annotation scheme to identify models and datasets associated with specific NLP tasks. Building on this \citet{dsouza-etal-2021-semeval} employed this annotation scheme to construct a knowledge graph that connects all models and datasets proposed for certain NLP tasks. Unlike these efforts, our work broadens the scope by identifying and categorizing contribution statements from NLP research papers without limiting them to specific NLP tasks. Additionally, our approach encompasses contributions that expand knowledge as well as introduce new artifacts.

\paragraph{Claim and Opinion Summarization.} Researchers have explored automated methods to study diverse perspectives of claims \citep{chen-etal-2019-seeing}. This includes the growing interest in key-point analysis \citep{bar-haim-etal-2021-every,friedman-etal-2021-overview}. Neural network and graph-based approaches have been proposed for claim summarization from newspaper reports and online discussions \citep{zhao-etal-2022-read, inacio-pardo-2021-semantic}. Some research has focused on extracting claims from %biomedical research
scientific  papers~\citep{achakulvisut2019claim,al-khatib-etal-2021-argument,sosa-etal-2023-detecting}.
While claims in research papers provide declarations to support hypotheses or research questions, contributions present new elements  (knowledge and artifacts) that a paper introduces to its field. In this work, we explore methods to extract and analyze contribution statements from NLP research papers. 

%% file: sections/03_dataset.tex
\section{\dataset{}: A Corpus of Contribution Statements}
\label{sec:data}

We developed a taxonomy of various types of contributions found in NLP research papers. Using this taxonomy, we annotated contribution statements from the abstracts of NLP research papers. We chose the abstracts as our corpus for annotation because abstracts are uniquely positioned at the beginning of papers and typically contain contribution statements. Moreover, abstracts efficiently summarize the paper, providing the context for understanding contributions, making them particularly suitable text segments to focus on for contribution annotation. Annotating entire papers would substantially escalate annotation efforts. %rendering 
Thus working with abstracts was a more efficient and effective option~\citep{teufel1999argumentative}.

\subsection{Taxonomy of Contributions}
\label{subsec:taxonomy}
In NLP research, contributions can broadly be divided into {\it two main types}. We call the first type {\it artifacts}, which encompasses the development of new or novel resources.
%such as datasets, models, or algorithms. 
NLP research heavily utilizes tools from machine learning, which relies on resources such as new methods or models, datasets - and the novel tasks they enable - all of which are recognized as significant contributions. Consequently, we categorize artifact contributions into three sub-types: {\it new methods} (a-methods), {\it new datasets} (a-datasets), and {\it new tasks} (a-tasks), each distinguished by the specific resource it brings to the field. 

%{\it models}, {\it datasets}, {\it methods} (or algorithms), and {\it tasks}, each distinguished by the specific resource it brings to the field. 

We term the second category as {\it knowledge} contributions that enrich the field with new insights or knowledge. Depending on what these contributions add knowledge to, we further categorize them
% Following the area to which these contributions add knowledge, these contributions are further categorized 
into five sub-types: {\it knowledge about method} (k-method), {\it knowledge about dataset} (k-dataset), {\it knowledge about task} (k-task), {\it knowledge about language} (k-language), and {\it knowledge about people} (k-people). This sub-categorization also mirrors the important elements in NLP research.

In Table~\ref{tab:annot_scheme}, we provide a detailed description of each type and subtype, with examples of contribution statements from research papers. While we acknowledge that alternative taxonomies may also be possible, we note that our proposed taxonomy is in line with the ACL'23 call for papers\footnote{\url{https://tinyurl.com/mpdkmzkj}}, which seeks submissions either that conduct analysis (thereby adding {\it knowledge}) or introduce new resources (thereby adding {\it artifacts}). 

% Additionally, we note that our taxonomy aligns with the ACL'23 call for papers\footnote{\url{https://2023.aclweb.org/calls/main_conference/}}, which seeks submissions either that conduct analysis (thereby adding {\it knowledge}) or introduce new resources (thereby adding {\it artifacts}). 
% However, it lacks granularity and fails to represent the diverse aspects of NLP, hence limiting its utility in examining the field's development. Additionally, while some research on NLP entity extraction \citep{hou-etal-2021-tdmsci} addresses entities similar to our defined artifact sub-types, it typically does not focus on contributions nor on the field's boundaries.

% \begin{figure}
%     \centering
%     \scalebox{0.80}{
%     %\includegraphics[width=0.45\textwidth]{asssets/mult_interpret_heatmap_v2.pdf}
%     \includegraphics[width=0.45\textwidth]{asssets/mult_interpret_heatmap_v3.pdf}
%     }
%     %\vspace*{-2mm}
%     \caption{Pointwise Mutual Information (PMI) between labels shows the co-occurrence of multiple labels in contribution statements.}
%     \label{fig:mult_interpret_hmap}
%     % \vspace*{-3mm}
% \end{figure}

\subsection{Curation}
\label{subsec:curation}

%\paragraph{Data Preparation.} 
\noindent\textbf{Data Preparation.} We compile a corpus of abstracts from 1,995 papers published under ACL Anthology using the S2ORC~\citep{lo2019s2orc}, a large collection of papers released to support research. We randomly selected these papers, guaranteeing a selection of at least five papers from each year between 1974 and February 2024. The selected papers were published in journals and conferences affiliated with ``ACL Events'', while those from workshops were excluded. %all published 
%at ``ACL Events'' that were journals and conferences. (We excluded papers from workshops.)
% under the ``ACL Events'' category. excluding those from workshops to maintain higher data quality. 
Additionally, we retrieve the metadata (i.e., unique id, title, authors, publication venue, and date) for each selected paper from \texttt{anthology.bib}.\footnote{\url{https://aclanthology.org/anthology.bib.gz}}

\begin{table}[t]
    \centering
    \scalebox{1.00}{
        \begin{tabular}{l l c}
        \toprule
        {\bf Typ.} & {\bf Sub-typ.} &  {\bf $\kappa$} \\
        \midrule
        \multirow{5}{*}{Knowledge} & k-dataset &  0.70 \\
        & k-language &  0.69 \\
        & k-method &  0.71 \\
        & k-people &  0.67 \\
        & k-task &  0.70 \\
        \midrule 
        %\multirow{3}{*}{Artifact} & Task & 3.6 & 0.76 \\
        \multirow{3}{*}{Artifact} & a-dataset &  0.76 \\        
        & a-method &  0.71 \\
        & a-task &  0.73 \\
        \midrule
        Overall Agreement & &  0.71 \\

        \bottomrule
        
        \end{tabular}
    }
    \caption{Inter-annotator Agreement}
    \label{tab:iaa}
    % \vspace*{-2mm}
\end{table}

\noindent{\textbf{Annotation.}} The main annotator is one of the authors of this paper, who has six years of experience in NLP research. Additionally, a PhD student with four years of research experience took part in annotation. We develop an annotation scheme to identify and classify the contribution statements in NLP research papers.
%, adhering to the previously proposed taxonomy. 
We use ontology-oriented annotation guidelines (refer to Appendix~\ref{app:annot_guidelines} for details) following \citet{liakata-etal-2010-corpora}. Regular meetings were conducted between the annotators to refine the guidelines as necessary~\citep{klie2024analyzing}.
%, following the recommendations by \citet{klie2024analyzing}.

Both annotators annotated 100 papers, ensuring representation from each decade between 1980 and 2024. We assess annotator agreements on these 100 papers. Subsequently, the senior annotator proceeded to annotate the abstracts of the additional 1,895 papers, adhering to the guidelines. Finally, the senior authors of this paper reviewed the dataset, particularly focusing on samples with disagreements, as a final check to ensure its quality.
We name the corpus of 1,995 annotated papers \dataset{}.

\noindent{\textbf{Agreement.}} We measure the inter-annotator agreement (IAA) by comparing the contribution statements from the 100 aforementioned papers annotated by the two annotators under the same contribution labels. All annotations were conducted in Label Studio~\citep{Label_Studio}. Table~\ref{tab:iaa} shows an average Fleiss' $\kappa$ of $0.71$, comparable to similar works on scholarly documents \citep{yang-li-2018-scidtb,hou-etal-2021-tdmsci,lauscher-etal-2022-multicite}. Further, we observe the error bounds of $\kappa$ between 0.60 (lower bound) and 0.82 (upper bound) with 95$\%$ confidence level ($p<0.05$). 

\subsection{Data Statistics}

We highlight three aspects of our dataset: first, it includes abstracts and metadata of $1,995$ papers from ``ACL Events'' in the ACL Anthology.  
%(with an average of $2.95$ sentences per abstract annotated as contribution statements; on average, an abstract consists of $5.42$ sentences), totaling $5,890$ annotated contribution statements. 
On average, each abstract comprises 5.42 sentences, with 2.95 sentences annotated as contribution statements, resulting in a total of 5,890 annotated contribution statements. 
Second, we illustrate the distribution of labels across these statements in Table~\ref{tab:lbl_dist}. Lastly, we note that $57.6\%$ of the contribution statements received multiple labels.  
% In Figure~\ref{fig:mult_interpret_hmap}, we illustrate the co-occurrence of different labels through Pointwise Mutual Information (PMI) scores. We observe a high co-occurrence between the labels {\it k-people} and {\it k-task}, likely because authors often explain how NLP tasks yield insights into humans or society. 
Figure~\ref{fig:mult_interpret_hmap} shows the co-occurrence of different contribution types within the same contribution statements, measured using pointwise mutual information (PMI) scores. Overall, the PMI values between any pair of contribution types are low, indicating low co-occurrence. However, {\it k-people} and {\it k-task} appear together more frequently than others, possibly because authors often explain how NLP tasks yield insights into humans or society. We divide our dataset into train-val-test (70-15-15) split at the paper level to maintain consistency in our experiments and prevent information leakage.

%% file: sections/04_exp.tex
\section{Automatically Identifying Contribution Statements and Contribution Types}
%\section{\dataset{} Evaluation}
%\section{\texttt{contribution} Context Classification}
%\section{Automated Classification of Contribution Statements}
%\section{Experiments}
\label{sec:exp}

\begin{figure}
    \centering
    \scalebox{0.80}{
    \includegraphics[width=0.6\textwidth]{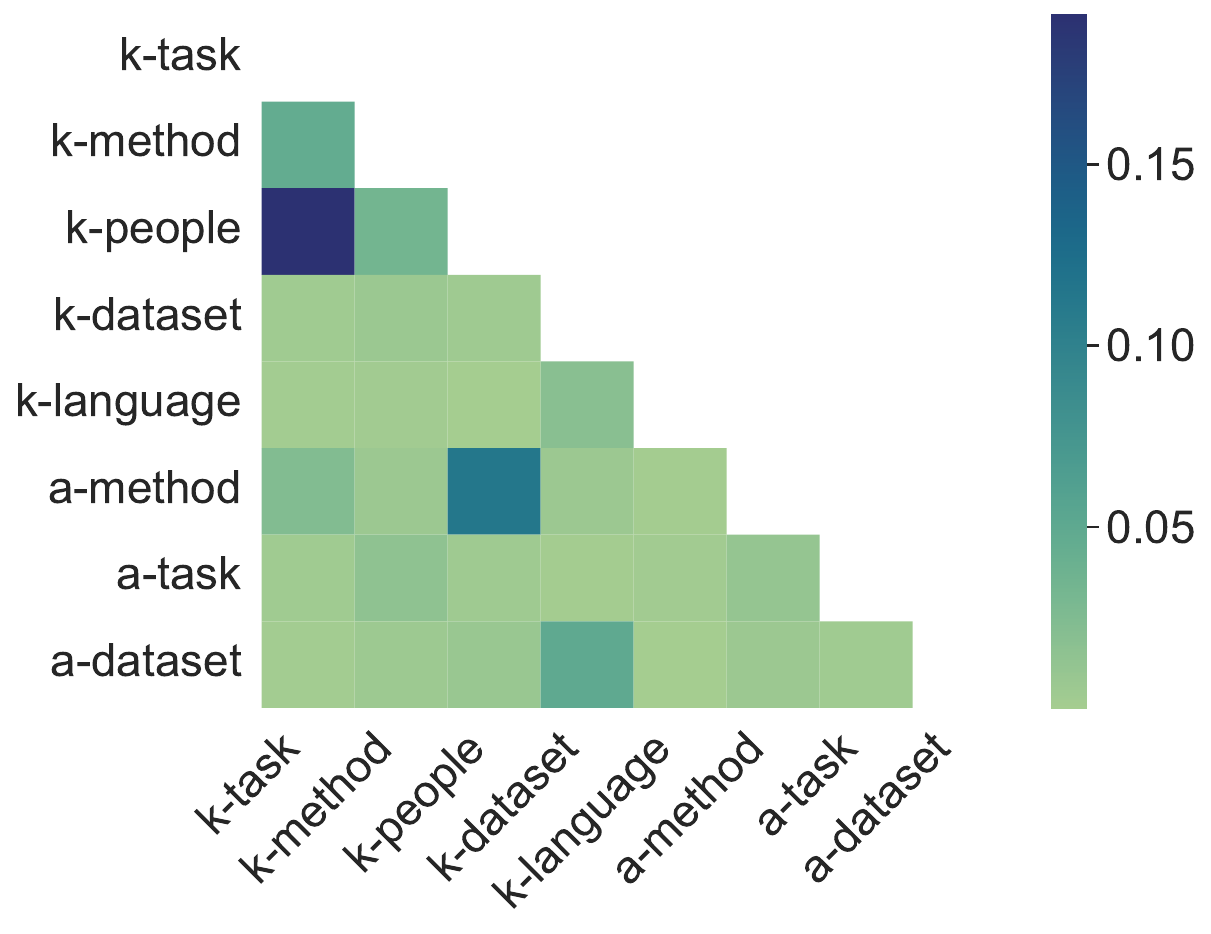}
    }
    %\vspace*{-2mm}
    % \caption{Pointwise Mutual Information (PMI) between labels shows the co-occurrence of multiple labels in contribution statements.}
    \caption{Pointwise mutual information (PMI) between contribution types shows the co-occurrence of multiple contribution types within the same contribution statements.}
    \label{fig:mult_interpret_hmap}
    % \vspace*{-4mm}
\end{figure}

We introduce the novel task of automatically detecting and categorizing contribution statements from NLP research papers. We use \dataset{} and benchmark multiple models to evaluate their performance on this task. 

%\subsection{New Task: Identifying Contribution Statements and Contribution Types}
%Contribution Statement Classification}
\subsection{Task Definition}
\label{subsec:task}

The task involves two steps: detecting contribution statements and subsequently categorizing them by type. We model it end-to-end as a multi-label extension of multi-class classification, where, given a statement, the objective is to assign it types and subtypes if it qualifies as a contribution; otherwise, assign Null. Formally, given a statement $S$, and a set of $n$ labels $L = [l_1, l_2, ..., l_n]$, the task is to predict a subset of these labels $Y = [y_1, y_2, ..., y_n]$ associated with $S$, where $y_j=1$, if $l_j$ is associated with $S$, and $0$ otherwise.

\begin{table}[t]
    \centering
    \scalebox{1.0}{
        \begin{tabular}{l l r}
        \toprule
        {\bf Typ.} & {\bf Sub-typ.} & {\bf Prop. ($\%$)} \\
        \midrule
        \multirow{5}{*}{Knowledge} & k-dataset & 5.1 \\
        & k-language & 4.0 \\
        & k-method & 12.6 \\
        & k-people & 9.2 \\
        & k-task & 36.1 \\
        % & ML & 12.6 \\
        % & People & 9.2 \\
        % & Dataset & 5.1 \\
        % & Language & 4.0 \\
        \midrule 
        %\multirow{3}{*}{Artifact} & Task & 3.6 & 0.76 \\
        \multirow{3}{*}{Artifact} & a-dataset & 2.2 \\
        & a-method & 27.2 \\
        & a-task & 3.6 \\
        %& Task & \mybar{0.04} & 0.75 \\
        % & Dataset & 2.2 \\
        \bottomrule
        
        \end{tabular}
    }
    % \caption{The proportion %distribution 
    % of Labels in the Annotated Contribution Statements.}
    \caption{Occurrence percentages of different contribution types in contribution statements from paper abstracts in \dataset{}.}
    \label{tab:lbl_dist}
    % \vspace*{-5mm}
\end{table}

%\subsection{Automatical Methods}
\subsection{Methods}
In our study, we explore two methods. 
%and train them using \dataset{}.
%of models to evaluate our dataset on the task. 
The first method involves utilizing pre-trained language models (PLMs) that are further fine-tuned using the training split of \dataset{}. Second, we use large language models (LLMs) and utilize prompting techniques for our task (refer to Appendix~\ref{app:add_res} for the prompting details). 
%This demonstrates that our dataset is suitable for developing contemporary NLP techniques.
We use the {\it binary relevance}~\citep{read2011classifier} for the task, treating each label as an independent binary classification problem. This avoids overfitting by not depending on previous label combinations and allows for flexible modifications to the label set without affecting other parts of the model. 
%Additionally, it prevents the risk of catastrophic forgetting during training.

\paragraph{PLMs.} We start our study with BERT~\citep{devlin-etal-2019-bert} and RoBERTa~\citep{liu2019roberta}, which are general-purpose pre-trained language models. Moving further, we use BiomedBERT~\citep{10.1145/3458754} and SciBERT~\citep{beltagy-etal-2019-scibert}, which are pre-trained on scientific texts. Additionally, we experiment with Flan-T5~\citep{chung2022scaling}, which is pre-trained over a collection of 1,836 fine-tuning tasks. We also implement a random baseline that assigns labels to sentences with a uniform random probability. 
%alongside a rule-based baseline following \citet{silvello2017semantic}.

\paragraph{LLMs.} Addressing the task through prompting, we use GPT-3.5-Turbo and GPT-4-Turbo \citep{openai2022gpt}, which are instruction-following large language models fine-tuned with reinforcement learning from human feedback (RLHF). Additionally, we use the open-source LLaMA-3-8B  model \citep{metaai2023llama}, which has been trained on over 15 trillion tokens gathered from publicly available domains.

\subsection{Training and Evaluation}
%For training the models, we divide our dataset into train-validation-test (70-15-15) split at the paper level to prevent information leakage. 
During fine-tuning the pre-trained language models, we use a grid search across various epochs $e\in \{1, 2, 3, 4, 5\}$ and learning rates $lr \in \{1\cdot 10^{-4}, 5\cdot 10^{-4}, 1\cdot 10^{-5}\}$, using a batch size of $32$. For prompting, we start with a zero-shot setting and gradually progress to a five-shot, respecting the context length limitations of the models. We repeat each experiment three times and observe the variance $<0.02$ for all of the models. 

Following \citet{uma2021learning}, for multi-label classification, we use label-based evaluation (macro-averaged precision, recall, and F1-score), which assesses performance on a per-label basis and then aggregates scores across all labels. We avoid label-set-based evaluation, also known as the exact match measure, because it does not effectively account for the sparsity characteristic of multi-labeling, often missing nuanced label variations.

% we evaluate our models using hard label evaluation (macro-averaged precision, recall, and F1-score). For multi-label classification, we use label-based evaluation, which assesses performance on a per-label basis and then aggregates scores across all labels. We avoid label-set-based evaluation, also known as the exact match measure, because it does not effectively account for the sparsity characteristic of multi-labeling, often missing nuanced label variations.

\subsection{Results and Discussion}
\label{subsec:benchmark}

\begin{table}[t]
    \centering
    \scalebox{0.9}{

        \begin{tabular}{l l c c c}
        \toprule
        {\bf Setting} & {\bf Model} & {\bf P} & {\bf R} & {\bf F1}\\
        \midrule
        & Random & 0.19 & 0.17 & 0.17 \\
        \midrule
        \multirow{5}{*}{Finetuning} & BERT & 0.31 & 0.50 & 0.38 \\
        & BiomedBERT & 0.64 & 0.59 & 0.60 \\
        & SciBERT & {\bf 0.81} & {\bf 0.80} & {\bf 0.80} \\
        & Flan-T5 & 0.79 & 0.78 & 0.78 \\
        & RoBERTa & 0.33 & 0.50 & 0.40 \\
        \midrule
        \multirow{3}{*}{Prompting} & GPT-3.5-Turbo & 0.75 & 0.71 & 0.73 \\
        & GPT-4-Turbo & 0.80 & {\bf 0.80} & 0.80 \\
        & LLaMA-3 & 0.60 & 0.56 & 0.53 \\
        \bottomrule
        
        \end{tabular}
    }
    \caption{Performance of different models for \texttt{contribution} statement classification.}
    \label{tab:eval_result}
    % \vspace*{-5mm}
\end{table}

Table~\ref{tab:eval_result} shows the results. We observe that SciBERT outperformed other fine-tuned pre-trained language models, likely due to its pre-training on a collection of scholarly documents. %which is closely aligned with our domain. 
Additionally, we note that GPT-4-Turbo's performance is on par with fine-tuned SciBERT. Hence, for environmental sustainability and cost-efficiency, we have chosen to use SciBERT for subsequent analyses. %However, 
Note that we tested the LLMs with five different prompt variants and recorded the most effective ones in Table~\ref{tab:llm_prompts} (Appendix~\ref{app:add_res}). We also found that LLM performance decreased when prompts included titles or entire abstracts, likely because titles may not accurately represent contributions, and LLMs are optimized for data with fixed context lengths. All reported results from experiments using pre-trained and large language models are statistically significant (McNemar's $p<0.001$). 
% We define two tasks on the \dataset{}.

\subsection{\datasetauto{}: A Corpus of Auto-Identified Contribution Statements}
%\subsection{\datasetauto{}: Automatically Identified Contribution Statements}
\label{subsec:larger_corpus}

We applied the fine-tuned SciBERT model to the sentences from the abstracts of papers in the ACL Anthology and classified them according to the predefined taxonomy. We call this corpus \datasetauto{}. This corpus can be used for diverse research purposes on NLP papers, including efficient semantic searches and key point analysis, among others. In the following section, we explore various NLP research trends using this corpus.

Specifically, we used S2ORC to gather the abstracts of $28,937$ papers published from conferences or journals falling within the ``ACL Events'' category between 1974 and February 2024 (details in Appendix~\ref{app:corpus_details}). We collected the metadata of these papers from the ``anthology.bib''. However, it is important to note here that while NLP papers are published outside of the ACL Anthology, it remains the largest single-source collection of NLP papers. Additionally, the Anthology's strict peer review process ensures high quality, making it a reliable source for our study.

%% file: sections/05_analysis.tex
\section{Analyzing the Nature of NLP}
\label{sec:analysis}

We study the nature of NLP by examining the trends and evolution in research contributions (\S~\ref{subsec:cont_and_evol}), the influence of publication venues (\S~\ref{subsec:venue_and_evol}), and their impact on citation patterns (\S~\ref{subsec:nlp_impact}). 

% We examine trends in NLP research contributions (\S~\ref{subsec:cont_and_evol}), their influence on venues (\S~\ref{subsec:venue_and_evol}), and citation counts (\S~\ref{subsec:nlp_impact}).

% We use the SciBERT model fine-tuned on \dataset{} to analyze and categorize contributions within the abstracts of 29,010 papers from the ACL Anthology. In this section, we will examine trends in NLP research contributions (\S~\ref{subsec:cont_and_evol}) and their potential future impact (\S~\ref{subsec:nlp_impact}). We will also discuss the relevance of our dataset, given the ongoing evolution of NLP research (\S~\ref{subsec:nlp_trends_analysis}), and explore how it can be utilized to track the development of specific topics within NLP.

\input{RQs/contribution_trends}

\input{RQs/contributions_venues}
\input{RQs/future_impact}

%% file: RQs/contribution_trends.tex
% \subsection{Contributions in NLP Research Papers and Their Evolution}
%\subsection{NLP Research Contributions Over Time}
\subsection{Evolving Contributions in NLP Research}
\label{subsec:cont_and_evol}

% The ACL Anthology is recognized as one of the largest collections of publications for NLP researchers, accurately reflecting the trends within the field. Following, \citet{hall-etal-2008-studying, tan-etal-2017-friendships}, we analyze papers from the ACL Anthology to discern the trends of contributions that advance NLP research. 
%\begin{compactenum}[start=1,label={Q\arabic*.}]
% \noindent{\it\textbf{Q1. What is the distribution of different contribution types in NLP research?}}
% \begin{enumerate}[start=1,label={Q\arabic*.}]
% \item What is the distribution of different contribution types in NLP research? 
% \end{enumerate}
% \noindent{\it\textbf{Q1. What is the distribution of different contribution types in NLP research?}}
\begin{enumerate}[wide, noitemsep, labelindent=0pt, topsep = 0pt, partopsep = 0pt, series=outerlist, start=1, label={\bf Q\arabic*.}]
\item \textbf{How do the various types of contributions shape the landscape of NLP research?}
\end{enumerate}
% \noindent{\it\textbf{Q1. How do the various types of contributions shape the landscape of NLP Research?}}

\begin{figure}
    \centering
    \scalebox{0.9}{
    \includegraphics[width=0.5\textwidth]{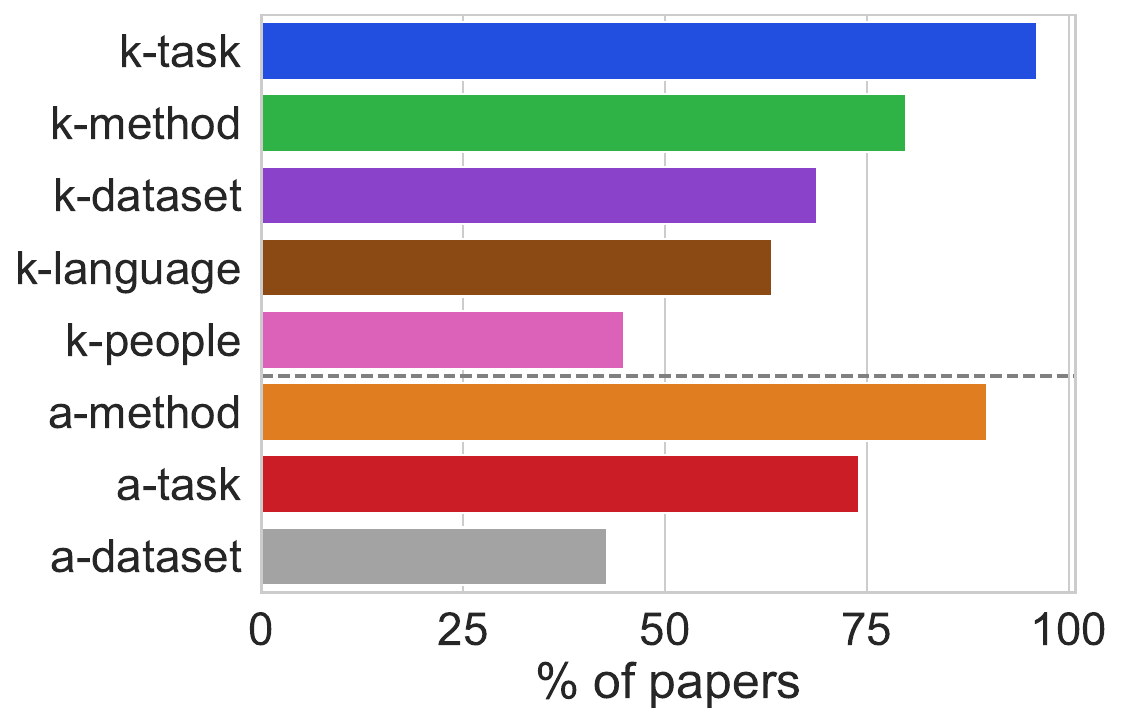}
    }
    % \caption{Distribution of contribution types across research papers.}
    \caption{Occurrence percentages of different contribution types associated with contribution statements in paper abstracts from \datasetauto{}.}
    \label{fig:k_a_cumulative_dist}
 %    \vspace*{-5mm}
\end{figure}

\noindent To study the breadth of contributions in NLP research, we examine the percentage of contribution statements associated with each contribution type and subtype in \datasetauto{}.
% \noindent To study the breadth of contributions in NLP research, we examine the distribution of contribution types in \datasetauto{}. 
% \noindent We study the breadth of contributions in NLP research, examining the distribution of contribution types in \datasetauto{}. 
% It is important to note that while NLP papers are published outside of ACL Anthology, it remains the largest single-source collection of NLP papers. Additionally, the Anthology's strict peer review process ensures high quality, making it a reliable source for our study.
%our corpus. 
% We identify their distinct types and sub-types in the abstracts of each paper and then aggregate these labels across the entire dataset. 
% Additionally, by reviewing the annual distribution of these labels, we study the development in the nature of contributions in NLP over time. 

%\noindent{\bf Results.} 
\paragraph{Results.}
Figure~\ref{fig:k_a_cumulative_dist} shows the distribution of different types of contributions in the abstracts of papers. 
Overall, we observe that:
% We illustrate the key results from Figure~\ref{fig:k_a_cumulative_dist} below. 
\begin{enumerate}[wide, noitemsep, labelindent=0pt, topsep = 0pt, partopsep = 0pt, label={\bf \alph*.}]
%\begin{enumerate}[wide, labelwidth=!, labelindent=0pt, label=\alph*)]
%    \item {\bf Fewer Contributions on people-knowledge.} Contributions towards {\it knowledge} about people ($44.9\%$) are less frequent compared to those concerning tasks ($89.8\%$) and methods ($78.7\%$).
\item Contributions of type {\it Knowledge about people} ($44.9\%$) 
and {\it Knowledge about language} ($61.2\%$) are relatively few
compared to contributions of type {\it k-task} ($89.8\%$) and {\it k-method} ($78.7\%$).
    \item % {\bf Prevalence of method contributions.} 
    Within the {\it artifact} type contributions, $\sim$$89\%$ of the papers introduce new methods (the highest), followed by tasks ($\sim$$75\%$) and, finally, datasets ($\sim$$45\%$).
\end{enumerate}

% that contributions towards {\it knowledge} about people ($44.9\%$) are less frequent compared to those concerning tasks and methods. Within the {\it artifact} type contributions, approximately $\sim89\%$ of the papers introduce new methods, the highest among artifact categories, followed by tasks and, finally, datasets. 

% \noindent{\bf Results.} Figure~\ref{fig:k_a_cumulative_dist} shows that contributions towards {\it knowledge} about people ($44.9\%$) are less frequent compared to those concerning tasks and methods. Within the {\it artifact} type contributions, approximately $\sim89\%$ of the papers introduce new methods, the highest among artifact categories, followed by tasks and, finally, datasets. 

%\noindent{\bf Discussion.}
\paragraph{Discussion.}
While some researchers suggest that NLP research is more relevant to people or society \citep{schober1992asking}, our observation reveals a significant focus of NLP research on knowledge about tasks and methods, particularly involving machine learning. Also, our findings resonate with those of \citet{pramanick-etal-2023-diachronic}, who noted through causal entity analysis that new methods and tasks have been drivers of NLP research. 
%\newline

% Figure~\ref{fig:k_a_cumulative_dist} shows regarding the contribution of type {\it knowledge}, that despite NLP's relevance to people or society \citep{schober1992asking}, only $44.9\%$ of papers contribute knowledge about people, while the majority of them contribute knowledge about tasks and ML. Contributions of type {\it artifact} reveal that $\sim89\%$ of the papers introduce new methods, the highest among artifact categories, followed by contributions to novel tasks and, finally, datasets. 

% These findings resonate with those of \citet{pramanick-etal-2023-diachronic}, who noted through causal entity analysis that new methods and tasks have been predominant drivers of NLP research since 1990.

% \noindent{\it\textbf{Q2. How has the distribution of contribution types in NLP research papers evolved over time?}}

\begin{enumerate}[wide, noitemsep, labelindent=0pt, topsep = 5pt, partopsep = 0pt, resume=outerlist, label={\bf Q\arabic*.}]
\item \textbf{How has the nature of NLP evolved over the years?}
\end{enumerate}
% \noindent{\it\textbf{Q2. How has the Nature of NLP evolved over the years?}}
% \begin{enumerate}[start=1,label={Q\arabic*.}, resume]
% \item How has the distribution of contribution types in NLP research papers evolved over time?
% \end{enumerate}

\begin{figure}
     \centering
     \begin{subfigure}[b]{0.5\textwidth}
         \centering
         \scalebox{0.95}{
         \includegraphics[width=\textwidth]{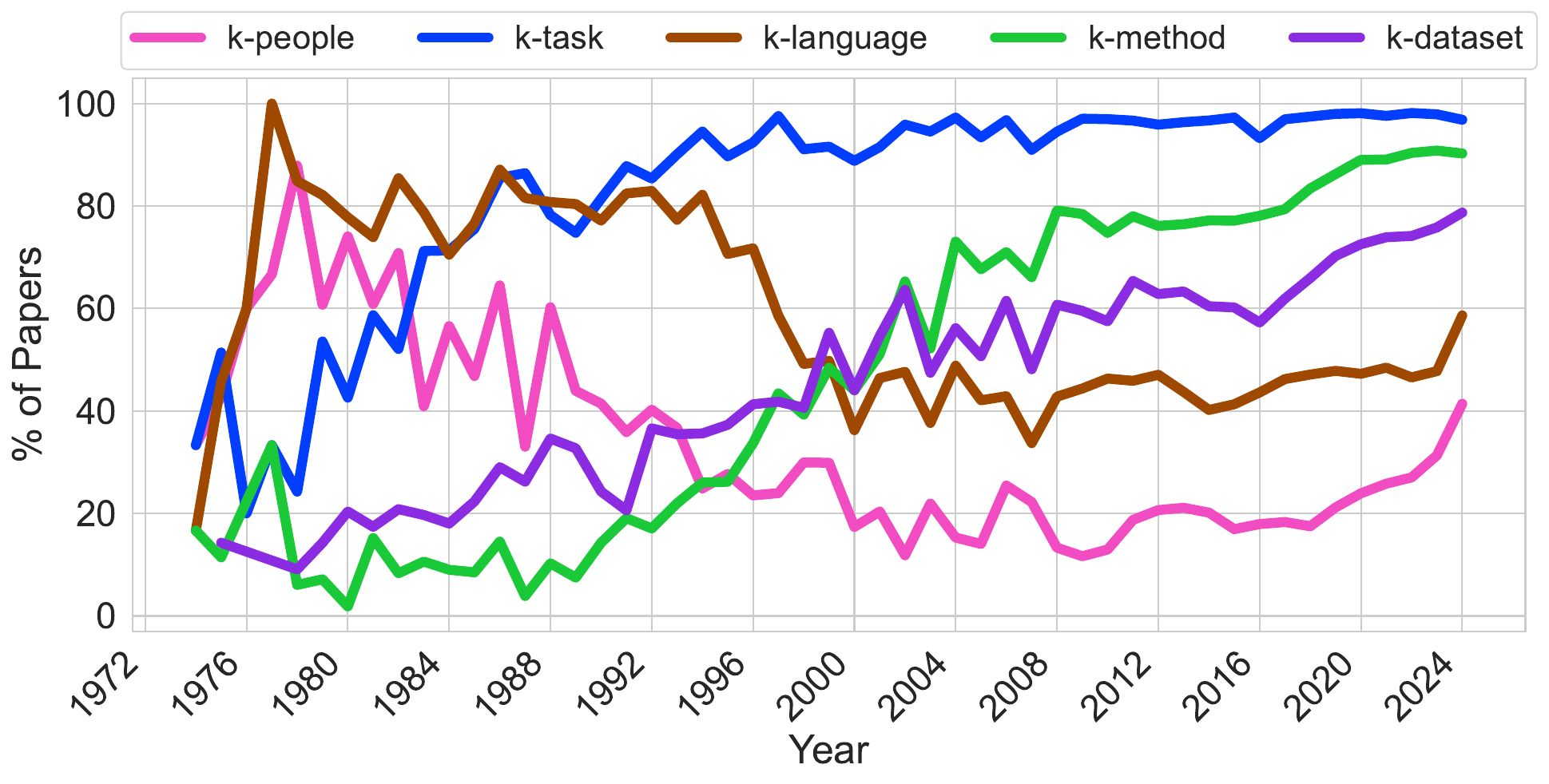}
         }
         \caption{Knowledge contributions}
         \label{subfig:temp_k_dist}
     \end{subfigure}
     \hfill
     %\hspace{1.5cm}
     \begin{subfigure}[b]{0.5\textwidth}
         \centering
         \scalebox{0.95}{
         % \includegraphics[width=\textwidth]{asssets/temporal_a_cumulative_label_dist.pdf}
         % }
         \includegraphics[width=\textwidth]{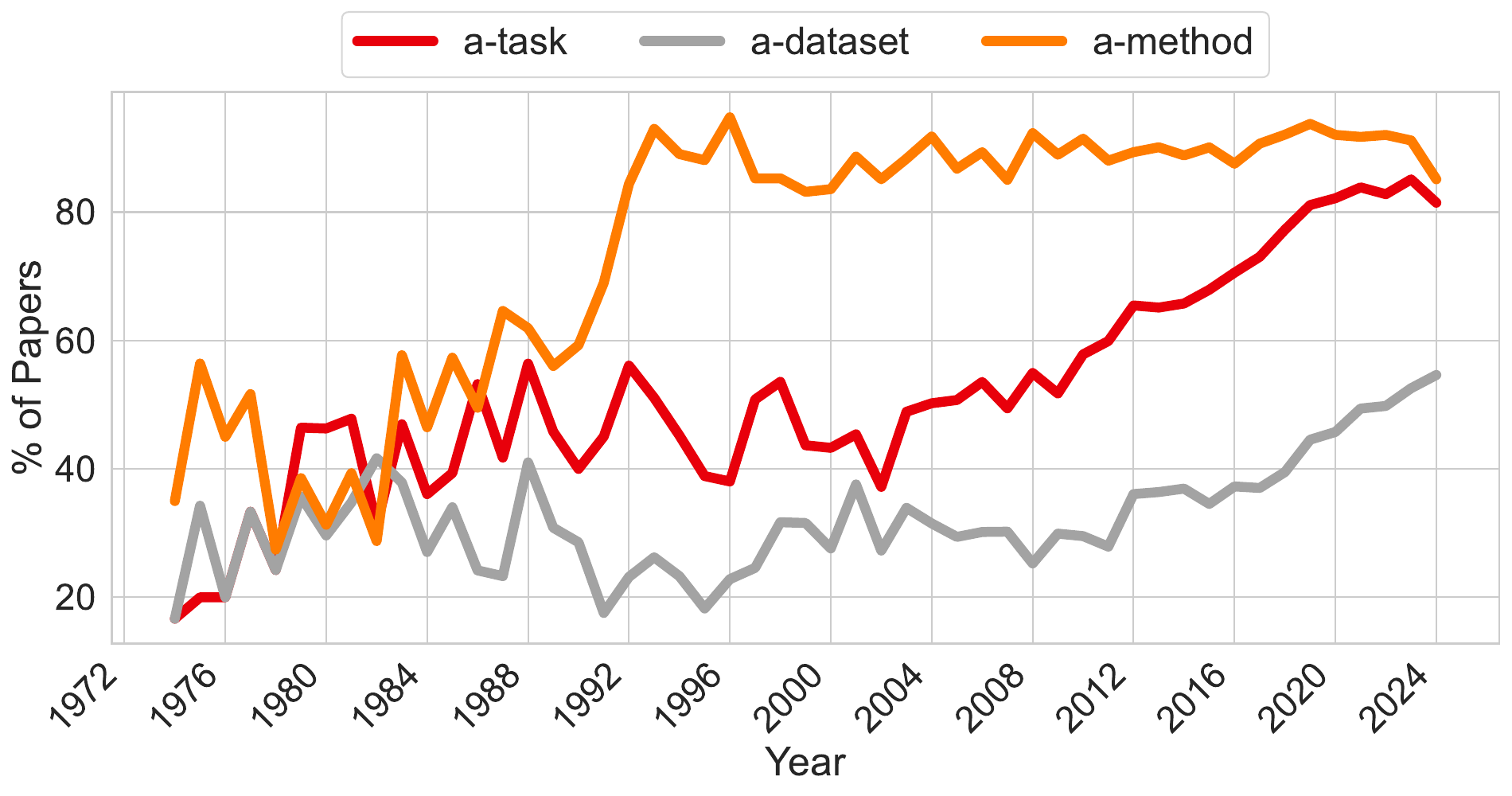}
         }
         \caption{Artifact contributions}
         \label{subfig:temp_a_dist}
     \end{subfigure}
     % \hfill
     % \begin{subfigure}[b]{0.3\textwidth}
     %     \centering
     %     \includegraphics[width=\textwidth]{graph3}
     %     \caption{$y=5/x$}
     %     \label{fig:five over x}
     % \end{subfigure}
        \caption{Percentage of papers in the \datasetauto{} that contain at least one contribution statement of various subtypes of (a) \textit{knowledge contributions} and (b) \textit{artifact contributions}.}
        % \caption{Temporal distribution of contribution types across research papers.}
        \label{fig:temp_dist_all}
        % \vspace*{-5mm}
\end{figure}
% \noindent To study how the nature of NLP research has changed over time, we examine the annual distribution of the contribution labels using the same methodology described in the previous question. 
\noindent To study how contribution types have evolved in NLP research, we calculate, for each year, the percentage of papers that include at least one contribution statement corresponding to each contribution type.
% \noindent We examine the annual distribution of the contribution labels using the same methodology described in the previous question.

%\noindent{\bf Results.} 
\paragraph{Results.}Figure~\ref{subfig:temp_k_dist} shows the following knowledge contribution trends:
\begin{enumerate}[wide, noitemsep, labelindent=0pt, topsep = 0pt, partopsep = 0pt, label={\bf \alph*.}]
\item {\bf Shift from language focus.} In the seventies and eighties, NLP focused on language, evidenced by significant contributions toward knowledge of language. However, from the early nineties to 2020, there was a dramatic decline in k-language contributions (drop to $\sim$$40\%$). Yet, post-2020, % there has been a resurgence of interest in and research 
we see a marked increase in contributions to language.  
\item {\bf Evolution in human-centric studies.} While early NLP research had a high percentage ($\sim$$80\%$) of k-people contributions, the percentage has declined steadily over time, dropping to $\sim$$20\%$ by the late 1990s. The percentage has stayed roughly steady since then until the late 2010s, when we see the beginnings of an increasing trend. 
% We observe a decline in contributions toward knowledge about people since the early eighties, indicating an early shift in NLP's focus. However, similar to the trends in language-focused research, contributions concerning people have seen a resurgence since post-2020. 

\item {\bf Consistent focus on NLP tasks.} During the early eighties, we observed a sharp increase in contributions focused on knowledge about NLP tasks, which has remained consistently high over the decades. 
% Since the early eighties, researchers recognized the importance of NLP tasks in research, leading to consistently high levels of contributions toward knowledge about NLP tasks throughout the decades.  

\item {\bf Steadily rising method and dataset knowledge.} Unlike other knowledge contributions, there has been a steady rise in contributions towards knowledge about datasets and methods over the years, with contributions to methods showing a more pronounced increase since the nineties.
\end{enumerate}

\noindent Figure~\ref{subfig:temp_a_dist} shows %illustrates 
artifact contribution trends as follows: 

\begin{enumerate}[wide, noitemsep, labelindent=0pt, topsep = 0pt, partopsep = 0pt, label={\bf \alph*.}]
\item {\bf Sharp rise in method artifacts.} We observe a sharp rise in contributions related to new methods beginning in the early nineties, which has sustained at high levels since then.

\item {\bf Steady rise in task and dataset artifacts.} Similar to methods, contributions toward artifact tasks have increased since the nineties. Both artifact tasks and datasets have shown a steady rise over the years, with a positive correlation between their growth.

\end{enumerate}

% \noindent{\bf Results.} Figure~\ref{subfig:temp_k_dist} shows a marked decline in contributions towards knowledge about language or people from the early nineties through 2020 (with a drop of $\sim$$40\%$). Figure~\ref{subfig:temp_a_dist} shows an increase in the number of papers proposing new methods during the same period. However, a resurgence in contributions towards knowledge about language and people is observed post-2020.  
% Around the same time, we also observed an increase in the intersection of NLP and machine learning. 

% we observe a marked increase in the intersection of NLP and machine learning starting around the year 1993. Since this time, we also observed a decline in contributions aimed at expanding knowledge about language or people. Figure~\ref{subfig:temp_a_dist} shows an increase in the number of papers proposing new methods during the same period. 

%\noindent{\bf Discussion.}
\paragraph{Discussion.}
In its early days (throughout the seventies and eighties), NLP research had a strong focus on language ~\citep{brachman-1979-taxonomy, lebowitz-1979-reading, herskovits-1980-spatial}. The early nineties marked a shift in NLP's focus with the advent of statistical models~\citep{brown1993mathematics}, the release of the Penn dataset~\citep{marcus-etal-1993-building}, and later the establishment of the EMNLP conference. While recent discussions often highlight newer methods or models (such as transformers or LLMs), our findings indicate that the shift towards contributions in methods or models began in the early nineties. That era set the stage for the development of newer methods that continue to shape NLP research. The recent rise in contributions toward new knowledge about people and language is likely due to the rise of new NLP sub-fields such as computational social science, culturonomics and digital humanities, and ethics in NLP (refer to Appendix~\ref{app_subsec:cont_and_evol} for an analysis of recent advancements in NLP research.). 
% an increased interest among researchers in sociolinguistics and the application of NLP in social sciences. 
% Lastly, it is important to recognize that quality datasets are important for NLP research, though their creation requires substantial effort and time. This is evident in the results, which show a gradual but steady increase in artifact contributions toward new datasets over the years. 
Finally, it is interesting to note that while different contribution types have ebbed and risen at different times, the percentages of \textit{all types} are moderate to high in the past five years. This is evidence that a great number of NLP papers are now contributing in multiple ways --- adding to knowledge and artifacts of many kinds (see supplementary analysis in Appendix~\ref{app_subsec:venue_and_evol}).

%% file: RQs/contributions_venues.tex
\subsection{Contributions and Venues}
\label{subsec:venue_and_evol}

% \begin{table}[!ht]
%     \centering
%     \scalebox{0.85}{
%     \begin{tabular}{l l c}
%     \toprule
%     & {\bf venue} & {avg. \#sent.} \\
%     \midrule
%     \multirow{8}{*}{conference} & ACL & 5.69 \\
%     & EMNLP & 6.23 \\
%     & NAACL & 5.74 \\
%     & EACL & 5.67 \\
%     & AACL & 6.43 \\
%     & Findings & 6.94 \\
%     & SEM & 5.48 \\
%     & CoNLL & 5.73 \\
%     \midrule
%     \multirow{2}{*}{journal} & TACL & 6.10 \\
%     & CL & 9.01\\
%     \bottomrule
%     \end{tabular}
%     }
%     \caption{Average number of sentences in abstracts for papers published across various conferences}
%     \label{tab:avg_abs_len}
% \end{table}

% Each publication venue maintains distinct expectations regarding the types of work it accepts, such as the focus on particular topics, the expected degree of polish, and the nature of the experiments conducted. Additionally, each venue often adheres to a unique genre of writing. 

% \noindent{\it\textbf{Q3. How do venues influence the types of contributions made in NLP research papers published at those venues?}}
\begin{enumerate}[wide, noitemsep, labelindent=0pt, topsep=0pt, partopsep=0pt, resume=outerlist, label={\bf Q\arabic*.}]
\item \textbf{How do venues influence the nature of NLP research?}
\end{enumerate}
%\noindent{\it\textbf{Q3. How do venues influence the nature of NLP Research?}}
% \begin{enumerate}[start=1,label={Q\arabic*.}, resume]
% \item How do venues influence the types of contributions made in NLP research papers published at those venues?
% \end{enumerate}

\noindent Each publication venue maintains distinct expectations regarding the types of work it accepts, such as the focus on particular topics or the nature of the experiments conducted. We examine the distinct types and further sub-types of the contribution statements in the abstracts of the papers across different venues normalized by the number of papers published in that venue. 

%\noindent{\bf Results.}
\paragraph{Results.}
We present detailed venue-specific statistics in Figure~\ref{fig:conf_dist} (Appendix~\ref{app:add_res}), and summarize the key findings below. 
% Below we summarize the key results.
\begin{enumerate}[wide, noitemsep, labelindent=0pt, topsep = 0pt, partopsep = 0pt, label={\bf \alph*.}]
\item {\bf Similar contributions across majority venues.} The majority of conferences (such as ACL, EMNLP, NAACL, etc.) display similar distributions regarding the types of contributions in their published papers: roughly $68\%$ artifacts (task: $71\%$, method: $89\%$, dataset: $42\%$) and $69\%$ knowledge (task: $94\%$, method: $77\%$, people: $44\%$, dataset: $65\%$, language: $61\%$).

\item {\bf Distinctiveness of EMNLP and CL.} EMNLP is distinguished by a notably higher volume of artifact-method contributions, highlighting its emphasis on empirical methodologies. Conversely, the CL journal is unique among *CL venues for its greater focus on expanding knowledge about language and people and a comparatively lesser emphasis on machine learning.
\end{enumerate}

\begin{enumerate}[wide, noitemsep, labelindent=0pt, topsep = 5pt, partopsep = 0pt, resume=outerlist, label={\bf Q\arabic*.}]
\item \textbf{How has the nature of NLP research papers changed across different venues over time?}
\end{enumerate}
% \noindent{\it\textbf{Q4. How has the nature of NLP Research papers changed across different venues over time?}}
% \begin{enumerate}[start=1,label={Q\arabic*.}, resume]
% \item How have the types of contributions in NLP research papers published at different venues changed over time?
% \end{enumerate}
\noindent We hypothesize that as a field matures, latent community norms develop, steering the research direction and leading to a more uniform distribution of contributions across different venues over time. To test this, we examine the change of specific types and sub-types of contributions in the abstracts of papers from each venue over time.

%\noindent{\bf Results.} 
\paragraph{Results.}
% Among the venues we analyze, CL is the oldest journal, and ACL is the oldest conference. 
We present the temporal distribution of contribution types across venues in Figure~\ref{fig:temp_conf_dist} (Appendix~\ref{app:add_res}) and summarize the key results below.

\begin{enumerate}[wide, noitemsep, labelindent=0pt, topsep = 0pt, partopsep = 0pt, label={\bf \alph*.}]

\item {\bf Spread of trends.} We first observe a decline in contributions concerning knowledge about people and language at the ACL in the early 1990s, which then gradually appears in CL.

\item {\bf Increasing similarity of newer conferences.} The trend towards similar contribution distributions is evident in newer conferences (such as EMNLP, NAACL, AACL, etc.). 
% The temporal distribution of contribution types across these venues is illustrated in Figure~\ref{fig:temp_conf_dist} (Appendix~\ref{app:add_res}).
\end{enumerate}

Additionally, in Appendix~\ref{app_subsec:venue_and_evol}, we examine whether these venues increasingly mirror ACL’s contribution distribution over time.

%% file: RQs/future_impact.tex
\subsection{Contributions and Citation Impact}
\label{subsec:nlp_impact}

% The narrative conveyed through research papers, particularly through the contributions they articulate, reinforces the technical competence of the work to the readers \citep{latour1987follow}. 
%These contributions shape how the work is perceived, which in turn influences how it is received and cited within the scholarly community \citep{shi2010citing}. 

% \noindent{\it\textbf{Q7.Are different types of research contributions associated with differing amounts of citation influence?}}
% \noindent{\it\textbf{Q7. Do different types of research contributions have differing amounts of citation influence?}}
\begin{enumerate}[wide, noitemsep, labelindent=0pt, topsep = 0pt, partopsep = 0pt, resume=outerlist, label={\bf Q\arabic*.}]
\item \textbf{How do different contribution types influence citation dynamics?}
\end{enumerate}
% \noindent{\it\textbf{Q7. How do different contribution types influence citation dynamics?}}
% \begin{enumerate}[start=1,label={Q\arabic*.}, resume]
% \item Are different types of research contributions associated with differing amounts of citation influence?
% \end{enumerate}

% \begin{table}[t]
%     \centering
%     \scalebox{0.65}{
%     \begin{tabular}{l l c c c}
%         \toprule
%         \multicolumn{2}{c}{Contribution} & \multirow{2}{*}{\#papers} & \multicolumn{2}{c}{\#citations} \\
        
%         % \multirow{2}{*}{avg. citation ($\uparrow$)} & \multirow{2}{*}{med. citation ($\uparrow$)}\\
%         % \cmidrule(lr){1-2}
%         {Typ.} & {Sub-typ.} & & mean & median\\
%         \midrule
%         \multirow{5}{*}{Knowledge} & k-dataset & 219 & 121.1 & 56.0 \\
%         & k-language & 193 & 107.1 & 53.0 \\
%         & k-method & 280 & 127.8 & 56.0 \\
%         %& Dataset & 219 & 121.1 & 56.0 \\
%         %& Task & 328 & 115.7 & 55.0 \\
%         & k-people & 119 & 109.5 & 54.0 \\
%         & k-task & 328 & 115.7 & 55.0 \\
%         %& Language & 193 & 107.1 & 53.0 \\

%         \midrule

%         \multirow{3}{*}{Artifact} & a-dataset & 154 & 137.7 & 64.0 \\
%         & a-method & 310 & 122.2 & 58.0 \\
%         & a-task & 270 & 116.0 & 56.0 \\

%         \bottomrule

%     \end{tabular}
%     }
    
%     \caption{Mean and median citation counts of papers for different contributions in ACL'18.}
%     \label{tab:reg_coeff}
%     \vspace*{-4mm}
% \end{table}

\noindent All contribution types are important for a thriving and vibrant ecosystem of NLP research. 
Thus, marked disparities in citation counts could potentially disincentivize work on certain types of contributions. Therefore, through this question, we track the citational impact of different contribution types.
% Hence, as a first step, it is important to analyze if differences exist in citation impact across various contribution types.
%The narrative conveyed through research papers often reinforces the technical competence of the work to the readers \citep{latour1987follow}. 
% Hence, in this analysis, we investigate the relationship between different contribution types and the number of citations a paper accrues. 
We calculate the average and median citation counts for each type of contribution from the papers that have at least one contribution statement pertaining to that type. 
%same type of contribution. 
To ensure a meaningful assessment of citation trajectories, we focus on papers with at least five years of publication history \citep{anderson2012towards}. For this purpose, we selected $352$ papers from the ACL'18 to examine the citation impact of papers published simultaneously for this experiment. 
%We aim to explore how these papers, published simultaneously, compare in terms of the citations they have received to date.

%\noindent{\bf Results.}
\paragraph{Results.}
Below we summarize the results presented in Table~\ref{tab:reg_coeff}. 

\begin{enumerate}[wide, noitemsep, labelindent=0pt, topsep = 0pt, partopsep = 0pt, label={\bf \alph*.}]

\item {\bf Dataset artifacts attract higher citations.} Regarding artifact contributions, papers that introduce new datasets tend to attract notably high citations. Additionally, those proposing new methods attract more citations compared to those introducing new tasks.

\item {\bf Greater interest in technical advancements.} Papers that contribute knowledge about methods or datasets (primarily through analysis) tend to receive more citations than those focused on people or language. This suggests a greater community interest in technical advancements over sociolinguistic studies.

\item {\bf Lower citation impact for language contributions.} Notably, even though more papers focus on expanding knowledge about language compared to those about people, language-focused contributions tend to receive fewer citations.

\end{enumerate}

We additionally analyze $277$ papers from ACL'17 (Table~\ref{tab:avg_citation_2017}, Appendix~\ref{app:add_res}), and our analysis reveals similar trends.

% \noindent{\bf Results.} Table~\ref{tab:reg_coeff} shows that the papers introducing new datasets receive notably high citations. Additionally, papers proposing new methods attract more citations than those proposing new tasks. Papers that add knowledge of methods or datasets (primarily through analysis) receive more citations than those that add knowledge about people or language, indicating the community's stronger interest in technical advancements over sociolinguistic studies. 
% %underscoring NLP's computational focus. 
% It is notable that even though there are more papers adding to the knowledge about language than with a focus on adding knowledge about people, the language-contribution papers tend to receive fewer citations.
% Further analysis of $277$ papers from ACL'17 (Table~\ref{tab:avg_citation_2017}, Appendix~\ref{app:add_res}) shows similar trends. % lending more reliability to our analysis. 

%\noindent{\bf Discussion.}
\paragraph{Discussion.}
It is important to recognize that citations are influenced by various factors beyond just contribution types. Our objective is neither to identify all possible influences on citation counts nor to pinpoint the most influential factors. 
% Conducting causal analysis in real-world settings is highly complex~\citep{elazar2024estimating} and falls outside the scope of this paper. 
However, the high citations for papers that create new datasets perhaps reflect the importance of datasets in much of NLP research, particularly for training and evaluating models - a common practice in modern NLP. Figure~\ref{app_fig:citation_boxplot} (Appendix~\ref{app:add_res}) shows the distribution of citation counts.

% Table~\ref{tab:reg_coeff} illustrates the association between specific types of contributions and the citation count of a paper. We find that contributions labeled as ``artifact:method;; and ``knowledge:ml'' show a higher correlation with citation impact, and all associations presented are statistically significant ($p<0.001$). It is important to note that citation rates are influenced by various factors. Our objective is neither to identify all possible influences on citation counts nor to pinpoint the most influential factors. Instead, we aim to determine which types of contributions are most closely associated with higher citation counts, among others.Further, we observe an $R^2$ value of $0.82$ to the regression fir. 

\begin{table}[t]
    \centering
    \scalebox{0.8}{
    \begin{tabular}{l l c c c}
        \toprule
        \multicolumn{2}{c}{Contribution} & \multirow{2}{*}{\#papers} & \multicolumn{2}{c}{\#citations} \\
        
        % \multirow{2}{*}{avg. citation ($\uparrow$)} & \multirow{2}{*}{med. citation ($\uparrow$)}\\
        % \cmidrule(lr){1-2}
        {Typ.} & {Sub-typ.} & & mean & median\\
        \midrule
        \multirow{5}{*}{Knowledge} & k-dataset & 219 & 121.1 & 56.0 \\
        & k-language & 193 & 107.1 & 53.0 \\
        & k-method & 280 & 127.8 & 56.0 \\
        %& Dataset & 219 & 121.1 & 56.0 \\
        %& Task & 328 & 115.7 & 55.0 \\
        & k-people & 119 & 109.5 & 54.0 \\
        & k-task & 328 & 115.7 & 55.0 \\
        %& Language & 193 & 107.1 & 53.0 \\

        \midrule

        \multirow{3}{*}{Artifact} & a-dataset & 154 & 137.7 & 64.0 \\
        & a-method & 310 & 122.2 & 58.0 \\
        & a-task & 270 & 116.0 & 56.0 \\

        \bottomrule

    \end{tabular}
    }
    
    \caption{Mean and median citation counts of papers for different contributions in ACL'18.}
    \label{tab:reg_coeff}
   % \vspace*{-3mm}
\end{table}

%% file: sections/06_conc.tex
\section{Conclusions and Discussion}
\label{sec:conc}

In this paper, we propose that automatically extracting, categorizing, and quantitatively analyzing contribution statements in research papers offers insights into the nature of the field. 
%We are the first to establish a taxonomy of contributions and develop a systematic framework to automatically extract, classify, and analyze contribution statements from NLP research papers (\S~\ref{sec:data}). 
We introduce a taxonomy of contributions and develop a framework for automatically processing the contribution statements from NLP papers (\S~\ref{sec:data}).

% This taxonomy and framework can be easily adapted to other fields with minimal expert intervention. 

Our analysis reveals that although NLP is intrinsically linked to linguistics and society, its current research focus is dominant towards advancements in technical methods (\S~\ref{subsec:cont_and_evol}). This shift toward newer methods, often discussed in the context of recent models like transformers and LLMs, actually began in the early nineties. However, an increased focus on methodology does not necessarily indicate a reduced emphasis on language or people. This is evident in post-2020 NLP research, where there is a growing interest in sociolinguistics and the use of NLP in social sciences alongside technical innovations like LLMs. Additionally, our analysis shows the field’s growth and progression, as reflected by the growing complexity and diversity of contribution types within research papers (\S~\ref{subsec:venue_and_evol}).

All contribution types play a vital role in sustaining a vibrant and dynamic NLP research ecosystem. Notably, we observe that artifact contributions -- particularly papers introducing new datasets -- tend to receive more citations than other types (\S~\ref{subsec:nlp_impact}). Although the growth of NLP is beneficial, we emphasize the importance of maintaining diversity in research contributions to ensure the field remains relevant to a broader community. As members of this community, we hold the strength to guide the future direction of these trends. However, we are not advocating for a specific stance on research practices but encourage an inclusive approach that embraces a variety of contribution types within NLP research. 
To foster future research in the area of contribution analysis and stimulate informed discussions within our community, we release our artifacts under \ccbyncsa.
%a non-commercial license. 
%\ccbyncsa.

%% file: sections/07_future.tex
\section{Applications and Future Work}

The \dataset{} dataset, which includes contribution statements annotated with their respective types, makes it valuable for a wide range of research projects and applications. Identifying contribution statements helps researchers efficiently navigate the growing body of literature by capturing the core ideas of each paper~\citep{fok-etal-2024-skimming}. The dataset, along with the proposed taxonomy, also holds promise for advancing tasks such as automatic survey generation~\citep{wang-etal-2024-autosurvey} and question answering within scientific literature~\citep{dasigi-etal-2021-dataset}. Categorizing contributions can further support researchers in locating studies with similar types of contributions or compiling structured literature review tables~\citep{newman-etal-2024-arxivdigestables}.

Beyond these applications, the dataset can be used to study how NLP research and its publication venues have evolved over time. For example, we are interested in studying the relationship between the diversity of contribution types and venue size, measured by the number of accepted papers and the range of research contributions represented. Additionally, we aim to explore and quantify the influence of different contributions of the same type, investigating how their impact evolves over time and what factors contribute to making a contribution influential. 

%% file: sections/limit.tex
\section*{Limitations}
\label{sec:limit}

This study primarily examines NLP research papers from the ACL Anthology, specifically focusing on papers from conferences and journals under ACL Events, such as ACL, EMNLP, NAACL, EACL, and journals like TACL and CL. However, it is crucial to recognize that significant NLP research also appears outside the ACL Anthology, including in AI venues, regional conferences, and preprint servers. While papers published in the ACL Anthology are typically of high quality, research from other venues often contributes valuable insights to the field. We leave the effort to curate and include research papers from these alternative venues for future work.

Our study primarily analyzes the abstracts of research papers, which are typically concise, logically coherent paragraphs that hold a unique position within the paper. While abstracts are likely to contain the key contributions as highlighted by the authors, making them a focal point for initial analysis, it is important to acknowledge that unique contributions may also be found within the main body of the paper. However, annotating the full text of research papers requires significant time, effort, and substantial domain knowledge to accurately understand and contextualize the content. In future iterations of this work, we plan to extend our annotations to include the main body of the papers, providing a more inclusive dataset.

Finally, for our analysis, we first train a classifier on the high-quality, human-annotated dataset that we create and then deploy this trained model on the larger ACL Anthology dataset. We conduct our analysis based on the labels generated by this model. It is important to acknowledge that no model achieves perfect accuracy, which can impact the quality of such analysis. However, as demonstrated by \citet{teodorescu2023evaluating}, when broader cumulative trends are derived from large datasets using such models, the results tend to be highly accurate and show a strong correlation with trends identified through gold-label analysis. This supports the reliability and accuracy of our analysis despite the inherent limitations of trained machine-learning models.

%% file: sections/ethics.tex
\section*{Ethics Statement}
\label{sec:ethics}

In this work, we utilize publicly accessible data from the ACL Anthology and do not involve any personal data. It is important to acknowledge that although our approach is data-driven, individual views on research are naturally subjective. Therefore, decisions in science should not only be based on data but also take into account ethical, social, and other qualitative considerations.

%% file: sections/ack.tex
\section*{Acknowledgements}

This work has been funded by the German Research Foundation (DFG) as part of the Research Training Group KRITIS No. GRK 2222. We also gratefully acknowledge Microsoft for providing access to OpenAI GPT models via the Azure cloud (Accelerate Foundation Model Academic Research).

We thank Aishik Mandal for his voluntary participation in the annotation study conducted for this research. We also appreciate the feedback on the initial draft of this manuscript provided by Hiba Arnaout, Sukannya Purkayastha, Fengyu Cai, Md Imbesat Hassan Rizvi, and Ilia Kuznetsov.

%% file: sections/appendix.tex
% \section{Example Appendix}
% \label{sec:appendix}

\input{app_sections/additional_analysis}

\input{app_sections/corpus_details}

\input{app_sections/annot_guide}

\input{app_sections/additional_results}

%% file: app_sections/additional_analysis.tex
\section{Supplementary Analysis}
\label{app:add_analysis}

In this section, we supplement the main analysis (\S~\ref{sec:analysis}) with additional insights to provide a comprehensive overview of the nature of NLP research.

% \subsection{Contributions in NLP Research Papers and Their Evolution}
% \subsection{Evolving Contributions in Contemporary NLP Research}
\subsection{Evolving Contributions in NLP Research}
\label{app_subsec:cont_and_evol}

\begin{enumerate}[wide, noitemsep, labelindent=0pt, topsep = 0pt, partopsep = 0pt, resume=outerlist, label={\bf Q\arabic*.}]
\item \textbf{How has the nature of NLP research evolved in recent years?}
\end{enumerate}

\noindent NLP research is experiencing an exciting phase~\citep{li-etal-2023-defining}, often referred to as the ``deep learning era''~\citep{pramanick-etal-2023-diachronic}, beginning in the late 2010s with the seminal work by \citet{vaswani2017attention}, followed by BERT~\citep{devlin-etal-2019-bert}, and the rise of Large Language Models (LLMs). We showed in Section~\ref{sec:analysis}, that the transformative shift in NLP research began in the early 1990s, setting the stage for these recent advancements. This section, however, focuses on the evolution of NLP research contributions since the late 2010s. 

% \noindent\paragraph{\bf Results.} 
\paragraph{Results.} 
\noindent We refer to Figure~\ref{subfig:temp_k_dist}, and summarize the key findings regarding knowledge contributions below.

\begin{enumerate}[wide, noitemsep, labelindent=0pt, topsep = 0pt, partopsep = 0pt, label={\bf \alph*.}]
\item {\bf Beginning of new research trends.} While contributions toward \textit{k-language} and \textit{k-people} declined in the early 1990s, the late 2010s (and especially early 2020s) have seen the beginning of a research trend marked by increased research contributions in these areas. 
\item {\bf Broad contributions spectrum.} In the last five years, there has been a moderate to high increase in the percentage of all types of contributions.
\end{enumerate}

We refer to Figure~\ref{subfig:temp_a_dist}, to summarize the key findings regarding artifact contributions. 

\begin{enumerate}[wide, noitemsep, labelindent=0pt, topsep = 0pt, partopsep = 0pt, label={\bf \alph*.}]
\item{\bf Steeper rise in dataset contributions.} Since the late 2010s, there has been a marked increase in research papers contributing new dataset artifacts, with \textit{a-dataset} showing a steeper rise compared to earlier periods.
\item{\bf Increasing new tasks.} Alongside the increasing contributions of type \textit{a-dataset}, there is also a rise in contributions toward \textit{a-task}, indicating a growth in the introduction of new tasks within NLP.
\end{enumerate}

% \noindent We refer to Figure~\ref{subfig:temp_k_dist}, and summarize the key findings regarding knowledge contributions below.

% \begin{enumerate}[wide, noitemsep, labelindent=0pt, topsep = 0pt, partopsep = 0pt, label={\bf \alph*.}]
% \item {\bf Beginning of new research trends.} While contributions toward k-language and k-people declined in the early 1990s, the late 2010s (and especially early 2020s) have seen the beginning of a research trend marked by increased research contributions in these areas. 
% \item {\bf Broad contributions spectrum.} In the last five years, there has been a moderate to high increase in the percentage of all types of contributions.
% \end{enumerate}

\paragraph{Discussion.} Newer models (such as the LLMs) excel at solving standard NLP tasks such as entity typing, sentiment analysis, and textual entailment~\citep{wei2021finetuned}. Rather than setting benchmarks, researchers explore new capabilities of these models and propose novel tasks~\citep{bubeck2023sparks}. These models are also adept at handling various complex tasks like chain-of-thought reasoning~\citep{wei2022emergent}. However, evaluating their capabilities often necessitates the collection of larger datasets, likely contributing to an increase in \textit{a-dataset} contributions. 

The increasing contributions to language knowledge may be tied to developments in newer models. Although Large Language Models (LLMs) are multilingual, i.e., trained on data from multiple languages, their performance is not uniformly effective across all languages for tasks such as classification or generation. To address this issue and improve model efficiency across various languages, researchers are increasingly focusing on studying the nuances of languages~\citep{aguilar2020lince}, thereby contributing to the knowledge of language. Similarly, efforts to address and mitigate social biases and stereotypes in LLM outputs~\citep{omrani2023social}, which often reveal their inherent flaws, have led to an increase in contributions focused on human-centered NLP.

\subsection{Contributions and Venues}
\label{app_subsec:venue_and_evol}

\begin{enumerate}[wide, noitemsep, labelindent=0pt, topsep = 0pt, partopsep = 0pt, resume=outerlist, label={\bf Q\arabic*.}]
\item \textbf{Are other venues mirroring the ACL conference in shaping the nature of NLP research?}
\end{enumerate}
% \noindent{\it\textbf{Q5. Are other venues mirroring the ACL Conference in shaping the nature of NLP research?}}
% \begin{enumerate}[start=1,label={Q\arabic*.}, resume]
% \item Are other venues increasingly resembling the ACL Conference in terms of NLP research contributions?
% \end{enumerate}

\begin{figure}
    \centering
    \scalebox{0.9}{
    \includegraphics[width=0.5\textwidth]{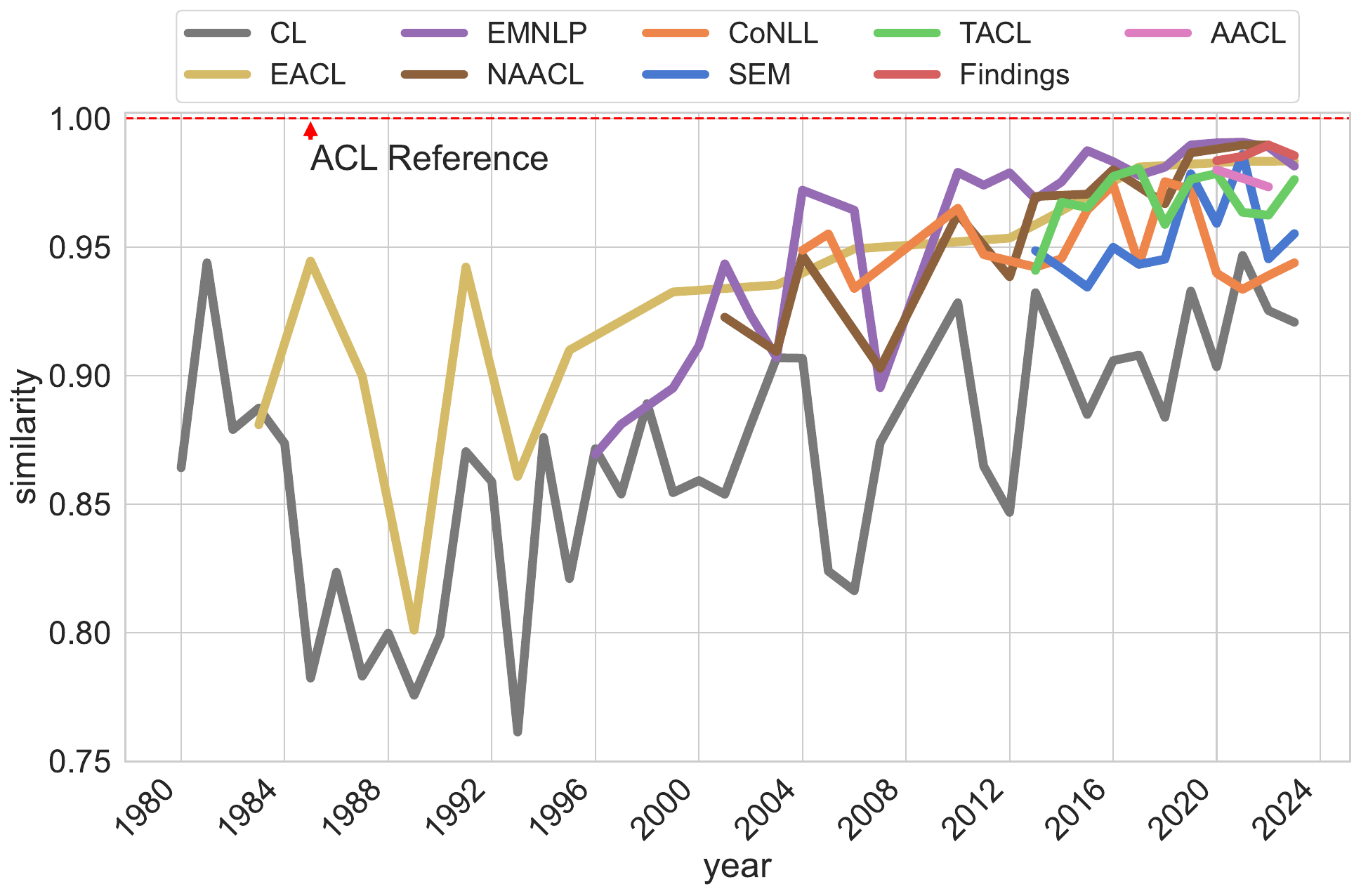}
    }
    \caption{Comparison of venue similarity based on contribution types.}
    \label{fig:conf_convergence}
    % \vspace*{-5mm}
\end{figure}

\noindent The ACL conference is the largest and arguably most prestigious among the ACL Events, hosting about $30.3\%$ of the papers published in these venues. Given its prominent position, it is interesting to investigate whether other conferences have gradually begun to mirror the distribution of the types of contributions featured in ACL over time. We compare the distribution of types and sub-types of contribution statements in papers from these venues with that of those from the ACL conference in the same year, using the Jensen-Shannon divergence~\citep{menendez1997jensen}, where a value close to $1$ indicates similar distributions.
% \noindent The growth of the NLP community has led to the introduction of new venues. Exploring the evolution of these new venues is particularly intriguing, especially whether they have become institutionalized to resemble established conferences or if they have developed unique stylistic identities that capture different aspects of knowledge. We analyze this by comparing the distribution of types and sub-types of contributions in papers from these venues with that of those from the ACL conference in the same year, using the Jensen-Shannon divergence~\citep{menendez1997jensen}, where a value close to $1$ indicates similar distributions.

\paragraph{Results.} Figure~\ref{fig:conf_convergence} shows the following trend across the venues. 

\noindent{\bf Convergence of NLP conferences with ACL.} Over the years, conferences have become increasingly similar to the ACL conference in terms of the distribution of the types of contributions their papers present. For instance, the EMNLP conference, originally established to focus on empirical findings, has shown growing similarity to ACL. Similarly, newer venues like AACL and Findings closely align with ACL’s contribution patterns.

\paragraph{Discussion.} This trend tends to confirm our hypothesis that, over time, a common publication norm has emerged across conferences, leading to a more institutionalized standard in NLP research. On the other hand, it is also arguably a loss that the different venues do not have unique characteristics, championing and valuing different kinds of works. \\

% matures and methods standardize, a common publication norm emerges across conferences, leading to a more institutionalized standard. This process is likely accelerated by repeated iterations of work within sub-communities, reinforcing these norms.

\begin{enumerate}[wide, noitemsep, labelindent=0pt, topsep = 0pt, partopsep = 0pt, resume=outerlist, label={\bf Q\arabic*.}]
\item \textbf{Do journal papers exhibit a greater variety of contribution types than conference papers?}
\end{enumerate}
% \noindent{\it\textbf{Q6. Do journal papers exhibit a greater variety of contribution types than conference papers?}}
% \begin{enumerate}[start=1,label={Q\arabic*.}, resume]
% \item Do journal papers exhibit a greater variety of contribution types compared to conference papers?
% \end{enumerate}

% \begin{figure}
%     \centering
%     \scalebox{0.85}{
%     \includegraphics[width=0.5\textwidth]{asssets/conf_avg_unique_contrib.pdf}
%     }
%     \caption{Average no. of contribution types per paper across different venues.}
%     \label{fig:conf_contrib}
%      %\vspace*{-3mm}
% \end{figure}

\noindent Different publication venues have varying constraints on the number of pages they allow, with journals typically offering more space compared to the stricter page limits at NLP conferences. To investigate whether journal papers utilize the additional space to include a wider range of contribution types, we analyze the average number of unique contributions per paper across various venues on an annual basis.

\begin{figure}
    \centering
    \scalebox{0.9}{
    \includegraphics[width=0.5\textwidth]{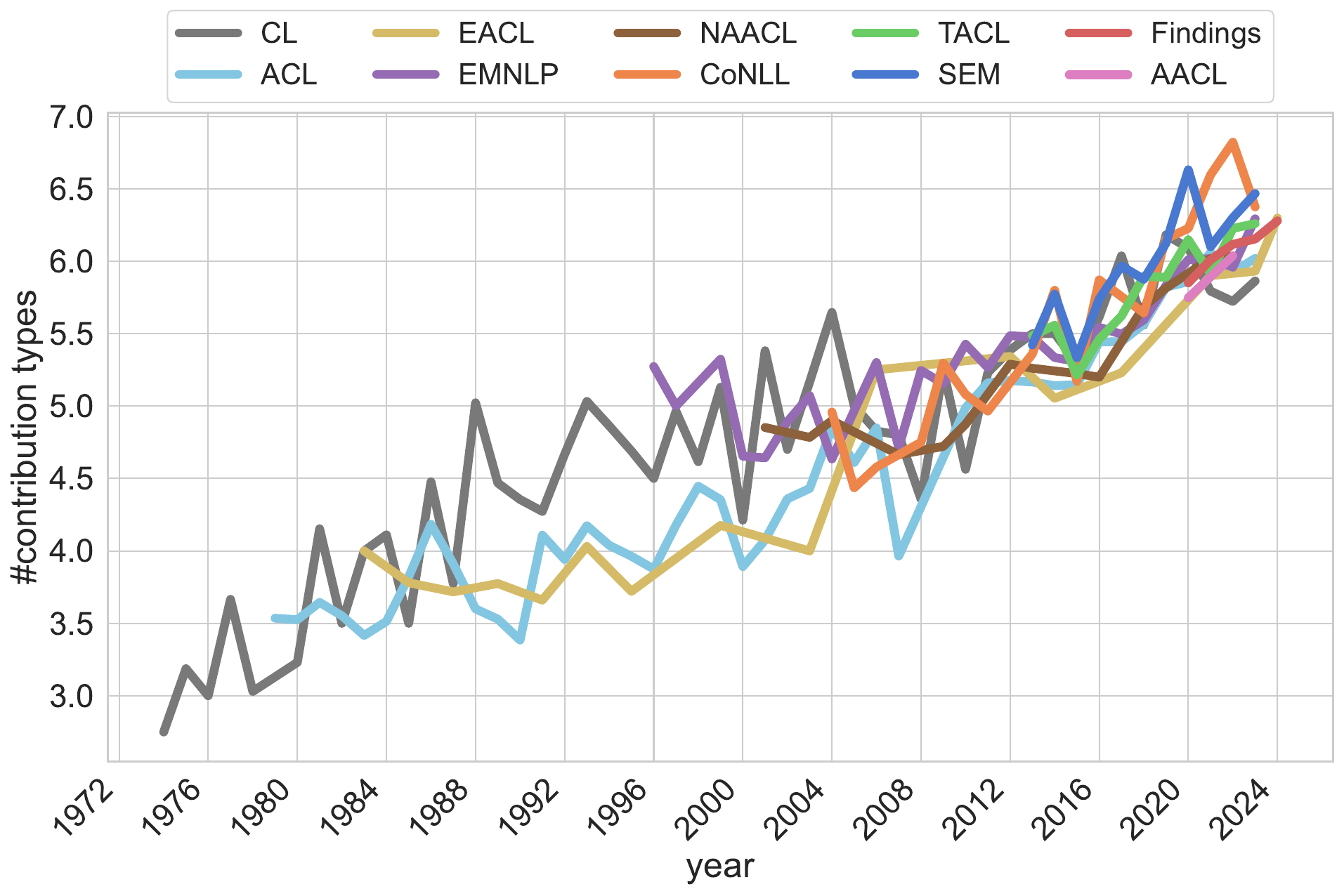}
    }
    \caption{Average number of contribution types per paper across different venues.}
    \label{fig:conf_contrib}
    % \vspace*{-5mm}
     %\vspace*{-3mm}
\end{figure}

\paragraph{Results.} Figure~\ref{fig:conf_contrib} shows the following.

\begin{enumerate}[wide, noitemsep, labelindent=0pt, topsep = 0pt, partopsep = 0pt, label={\bf \alph*.}]
\item {\bf Rising diversity in contributions.} The average number of unique contribution types in the abstracts of conferences and journals has been consistent, and this number has shown an upward trend over time. 
\item {\bf Expansion in NLP applications.} The consistent average length of abstracts in venues, yet diverse contribution types (Figure~\ref{fig:conf_abs}, Appendix~\ref{app:add_res}) indicates the growing sophistication and expansion of NLP applications.
\end{enumerate}

% \noindent{\bf Results.} Figure~\ref{fig:conf_contrib} shows that conferences and journals have similar average number of unique contribution types in their abstracts, but this number has been increasing over time.
% %, suggesting that NLP research papers include more diverse contribution types. 
% However, we observe that the average length of abstracts in venues has not changed significantly over time (Figure~\ref{fig:conf_abs}, Appendix~\ref{app:add_res}). The results indicate the growing sophistication and expansion of NLP applications. 

%% file: app_sections/corpus_details.tex
\section{Corpus Details}
\label{app:corpus_details}

To address the question {\it ``What constitutes NLP Research?''}, first 
We used S2ORC to gather the abstracts of $29,010$ papers published from conferences or journals falling within the ``ACL Events'' category. Notably, in 1997, the ACL and EACL conferences were held jointly, resulting in 73 papers being listed under both events for that year. We treated these dual-listed papers as single entries, reducing our dataset to a total of $28,937$ unique research papers. Additionally, we collected the metadata associated with these papers from the \texttt{anthology.bib}. 

Finally, we applied the SciBERT model, fine-tuned on \dataset{}, to the sentences from these abstracts to identify the contribution statements and classify them according to the predefined taxonomy.

%% file: app_sections/annot_guide.tex
\section{\dataset{} Annotation Guidelines}
\label{app:annot_guidelines}

% \begin{table}[!ht]
%     \centering
%     \scalebox{0.75}{
%         \begin{tabular}{l l c}
%         \toprule
%         {\bf Typ.} & {\bf Sub-typ.} & {\bf $\#$Contrib.} \\
%         \midrule
%         \multirow{5}{*}{Knowledge} & k-dataset & 25 \\
%         & k-language & 23 \\
%         & k-method & 53 \\
%         & k-people & 41 \\
%         & k-task & 122 \\
%         % & ML & 12.6 \\
%         % & People & 9.2 \\
%         % & Dataset & 5.1 \\
%         % & Language & 4.0 \\
%         \midrule 
%         %\multirow{3}{*}{Artifact} & Task & 3.6 & 0.76 \\
%         \multirow{3}{*}{Artifact} & a-dataset & 20 \\
%         & a-method & 94 \\
%         & a-task & 22 \\
%         %& Task & \mybar{0.04} & 0.75 \\
%         % & Dataset & 2.2 \\
%         \bottomrule
        
%         \end{tabular}
%     }
%     \caption{The number %distribution 
%     of Labels in the Annotated Contribution Statements by the two annotators.}
%     \label{tab:joint_annot_stat}
%     %\vspace*{-5mm}
% \end{table}

We propose a linguistic annotation scheme to study and analyze the types of contributions articulated in NLP research papers following the taxonomy we developed in \S~\ref{subsec:taxonomy}. The goal of this annotation scheme is to annotate contribution statements from the abstract section of NLP research papers into various types and further into sub-types as mentioned in Table~\ref{tab:annot_scheme}, according to the specific aspect of field advancement they represent. In this section, we elaborate on both broad categories and more detailed sub-types of contributions within each category.

\subsection{Contribution Types}
\label{app:contrib_types}

\subsubsection{Type: Artifact}

This type of contribution includes the creation of new resources such as datasets, models, or algorithms. We further sub-categorize contributions into three distinct categories based on the type of artifact they introduce to the field.

\begin{itemize}
    \item {\it a-dataset:} Researchers create new scientific corpus or language resources as artifacts to build models or analyze languages, such as SQuAD or Penn Treebank.
    \item {\it a-method:} Researchers often create new NLP methods such as algorithms or models (for example, BERT (pre-trained language model) or LLaMA (large language model)), as artifacts primarily to solve tasks and describe them in research papers. 
    \item {\it a-task:} Researchers often identify or formulate new or previously unknown problems (such as linguistic problems like NER Tagging) and formally describe them in research papers as tasks. 
    % \item {\it Models:} Researchers often create new NLP models such as BERT (pre-trained language model) or LLaMA (large language model) as artifacts to solve tasks and describe them in research papers. 
    % \item {\it Methods (or algorithms):} Researchers develop new methods or algorithms to develop models, analyze datasets, or even solve tasks and describe them in research papers.  
\end{itemize}

\subsubsection{Type: Knowledge}

This type of contribution encompasses the addition of new insights or understandings to the field. These contributions often relate to linguistic studies due to the significant overlap between NLP and linguistics, or they may explore societal and human aspects because of NLP’s focus on human language. Consequently, we further categorize it into five sub-types based on the specific area of knowledge it expands. 

\begin{itemize}
    \item {\it k-dataset:} Contribute new insights or analysis of an NLP dataset. 
    \item {\it k-language:} Adds new knowledge about natural language. 
    \item {\it k-method:} Enhances the understanding of algorithms, methods or models within NLP.
    \item {\it k-people:} Explores and adds knowledge about aspects of human behavior and social implications as revealed through natural language. 
    \item {\it k-task:} Contribute new insights into specific NLP task(s).
\end{itemize}

\subsection{Annotation Instructions}

First, we present the following two definitions to the annotators along with the contribution types as described in \S~\ref{app:contrib_types}. 

\begin{definition}[Contribution]
A contribution is a scientific achievement attributed to the authors of a research paper, such as introducing a new model or dataset. 
\end{definition}

\begin{definition}[Contribution Statement]
A statement in a research paper that describes new scientific achievements attributed to its authors is called a contribution statement. 
\end{definition}

Next, we present the annotators with the title and abstract of a research paper and instruct them to annotate each statement of the abstract according to the following questions. 

\begin{enumerate}[start=1,label={Q\arabic*.}]
\item {\it Does the sentence qualify as a {contribution statement} according to the definitions presented earlier?}
\end{enumerate}

In the second question, multiple options could be selected. 

\begin{enumerate}[label={Q\arabic*.}, resume]
\item {\it If you answered yes to the previous question, which of the following options most accurately describes the type and sub-type of the contribution statement? Please select all that apply.}
\end{enumerate}

\begin{itemize}
    \item {\it Artifact-Task} (Introduces, proposes, or formulates a new or novel NLP task.)
    %\item {\it Artifact-Model} (Builds or creates new NLP model.)
    \item {\it Artifact-Method} (Introduces or creates a new or novel NLP method such as an algorithm or a new NLP model.)
    \item {\it Artifact-Dataset} (Creates a new corpus or language resource.)
    % \item {\it Artifact-Method} (Introduces or proposes a new or novel NLP method or algorithm.)
    \item {\it Knowledge-Task} (Describes new knowledge about NLP task.)
    \item {\it Knowledge-Dataset} (Describes new knowledge about datasets, such as their new properties or characteristics.)
    \item {\it Knowledge-Method} (Describes or presents new knowledge or analysis about NLP models or methods, which are primarily drawn from Machine Learning.)
    \item {\it Knowledge-Language} (Presents new knowledge about language, such as a new property or characteristic of language.)
    \item {\it Knowledge-People} (Presents new knowledge about people, humankind, society, or human civilization.)
    \item {Others} (Any other type that does not fall under the categories mentioned above.)
\end{itemize}

\paragraph{Discussion:} We observe that only a small number of NLP papers propose new {\it metrics for evaluation measures}. Since these metrics function as algorithms or methods, we categorize them under the class ``artifact-method''. 
%Additionally, while new NLP methods or models could broadly be considered artifacts of Computer Science, we observe that NLP primarily utilizes Machine Learning (ML) tools to develop these methods or models. Therefore, we categorize them under the class ``artifact:ml''.

Also note that, in the initial pilot annotation study, we included an ``Others'' label (as mentioned above) to capture any types of contributions not already accounted for in our taxonomy. This allowed us to potentially expand our taxonomy based on the pilot results. Following the pilot study, however, we found that our existing taxonomy adequately covered all types of contributions identified by the annotators.

\subsection{Annotation Statistics}

\begin{table}[!ht]
    \centering
    \scalebox{0.75}{
        \begin{tabular}{l l c}
        \toprule
        {\bf Typ.} & {\bf Sub-typ.} & {\bf $\#$Contrib.} \\
        \midrule
        \multirow{5}{*}{Knowledge} & k-dataset & 25 \\
        & k-language & 23 \\
        & k-method & 53 \\
        & k-people & 41 \\
        & k-task & 122 \\
        % & ML & 12.6 \\
        % & People & 9.2 \\
        % & Dataset & 5.1 \\
        % & Language & 4.0 \\
        \midrule 
        %\multirow{3}{*}{Artifact} & Task & 3.6 & 0.76 \\
        \multirow{3}{*}{Artifact} & a-dataset & 20 \\
        & a-method & 94 \\
        & a-task & 22 \\
        %& Task & \mybar{0.04} & 0.75 \\
        % & Dataset & 2.2 \\
        \bottomrule
        
        \end{tabular}
    }
    % \caption{The number %distribution 
    % of contribution types in the Annotated Contribution Statements by the two annotators.}
    \caption{Number of annotated contribution statements pertaining to each contribution type, as annotated by two annotators.}
    \label{tab:joint_annot_stat}
    %\vspace*{-5mm}
\end{table}

We provide statistics for the abstracts of 100 papers annotated by two annotators post-adjudication. Of the 584 statements annotated, 359 were identified as contributions. Table~\ref{tab:joint_annot_stat} details the number of these statements across each type and subtype of contributions.

%% file: app_sections/additional_results.tex
\section{Supplementary Results}
\label{app:add_res}

\begin{table}[!ht]
    \centering
    \scalebox{0.9}{
        \begin{tabular}{l l c}
        \toprule
        {\bf Typ.} & {\bf Sub-typ.} & {\bf macro-F1}\\
        \midrule
        \multirow{5}{*}{Knowledge} & k-dataset & 0.80 \\
        & k-language & 0.80 \\
        & k-method & 0.80 \\
        & k-people & 0.81 \\
        & k-task & 0.81 \\
        \midrule
        \multirow{3}{*}{Artifact} & a-dataset & 0.80 \\
        & a-method & 0.81 \\
        & a-task & 0.80 \\
        \bottomrule
        
        \end{tabular}
    }
    \caption{SciBERT's performance in identifying statements for each contribution type.}
    \label{app_tab:lbl_eval_result}
\end{table}

\begin{table}[!ht]
    \centering
    \scalebox{0.85}{
    \begin{tabular}{l l c}
    \toprule
    & {\bf venue} & {avg. \#sent.} \\
    \midrule
    \multirow{8}{*}{conference} & ACL & 5.69 \\
    & EMNLP & 6.23 \\
    & NAACL & 5.74 \\
    & EACL & 5.67 \\
    & AACL & 6.43 \\
    & Findings & 6.94 \\
    & SEM & 5.48 \\
    & CoNLL & 5.73 \\
    \midrule
    \multirow{2}{*}{journal} & TACL & 6.10 \\
    & CL & 9.01\\
    \bottomrule
    \end{tabular}
    }
    \caption{Conference-wise average abstract sentence count.}
    % \caption{Average number of sentences in abstracts for papers published across various conferences.}
    \label{tab:avg_abs_len}
\end{table}

\begin{table}[!ht]
    \centering
    \scalebox{0.9}{
    \begin{tabular}{l c c c c}
    \toprule
    {\bf Model} & {0-shot} & {1-shot} & {3-shot} & {5-shot} \\
    \midrule
    GPT-3.5-Turbo & 0.52 & 0.55 &  0.64 &  0.73\\
    GPT-4-Turbo &{\bf 0.61} & {\bf 0.66} & {\bf 0.72} & {\bf 0.80} \\
    LLaMA-3 & 0.50 & 0.50 & 0.51 & 0.53 \\
    \bottomrule
    \end{tabular}
    }
    \caption{LLM performance (\texttt{macro-F1}) with different number of training examples.}
    \label{tab:llm_shots}
\end{table}

\begin{table}[!ht]
    \centering
    \scalebox{0.9}{
    \begin{tabular}{l l c c}
        \toprule
        \multicolumn{2}{c}{Contribution} & \multirow{2}{*}{\#papers} & \multirow{2}{*}{avg. citation ($\uparrow$)}\\
        \cmidrule(lr){1-2}
        {Typ.} & {Sub-typ.} & \\
        \midrule
        \multirow{5}{*}{Knowledge} & k-dataset & 180 & 120.0\\
        & k-language & 160 & 110.3\\
        & k-method & 230 & 135.3\\
        & k-people & 101 & 104.7\\
        & k-task & 254 & 121.5 \\
        %& ML & 230 & 135.3\\
        %& People & 101 & 104.7\\
        %& Dataset & 180 & 120.0\\
        %& Language & 160 & 110.3\\

        \midrule

        \multirow{3}{*}{Artifact} & a-dataset & 96 & 140.7 \\
        & a-method & 250 & 131.5 \\
        & a-task & 202 & 120.8\\
        %& Dataset & 96 & 140.7 \\
        
        \bottomrule

    \end{tabular}
    }
    
    \caption{Average citation counts by contributions for ACL 2017 Papers.}
    \label{tab:avg_citation_2017}
    %\vspace*{-3mm}
\end{table}

\begin{figure}
    \centering
    \scalebox{0.85}{
    \includegraphics[width=1\columnwidth, height=26.0cm]{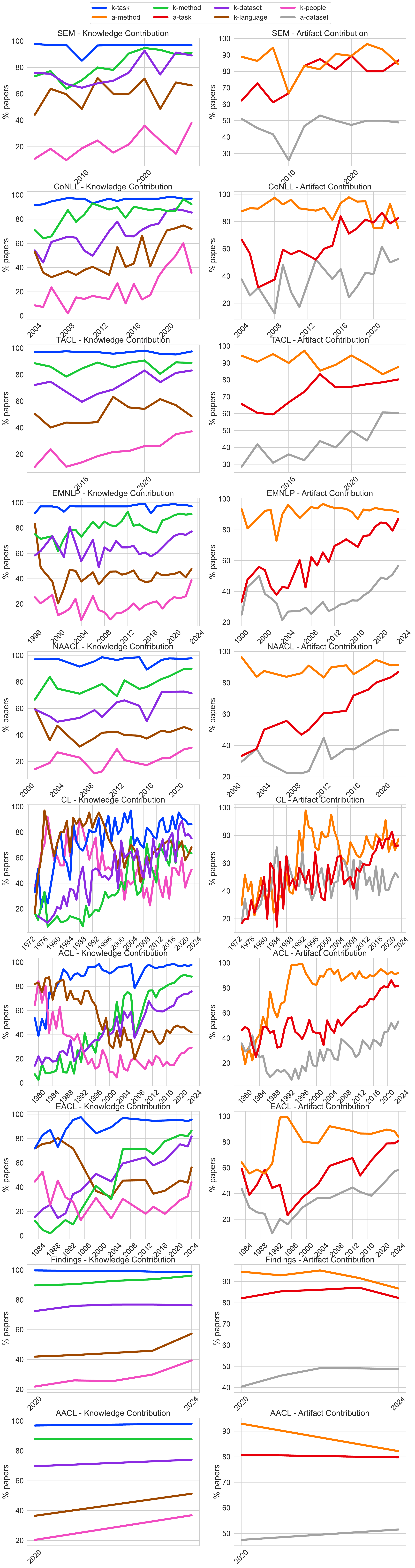}
    }
    \caption{Evolution of NLP conferences (and journals) based on the percentage of papers containing at least one contribution of each type (Abbr.: knowledge (k), artifact (a)). Refer to Figure~\ref{fig:temp_conf_dist_p1}, ~\ref{fig:temp_conf_dist_p2},~\ref{fig:temp_conf_dist_p3} for larger figures.}
    \label{fig:temp_conf_dist}
     %\vspace*{-3mm}
\end{figure}

\begin{figure*}
    \centering
    \scalebox{1}{
    \includegraphics[width=1\textwidth]{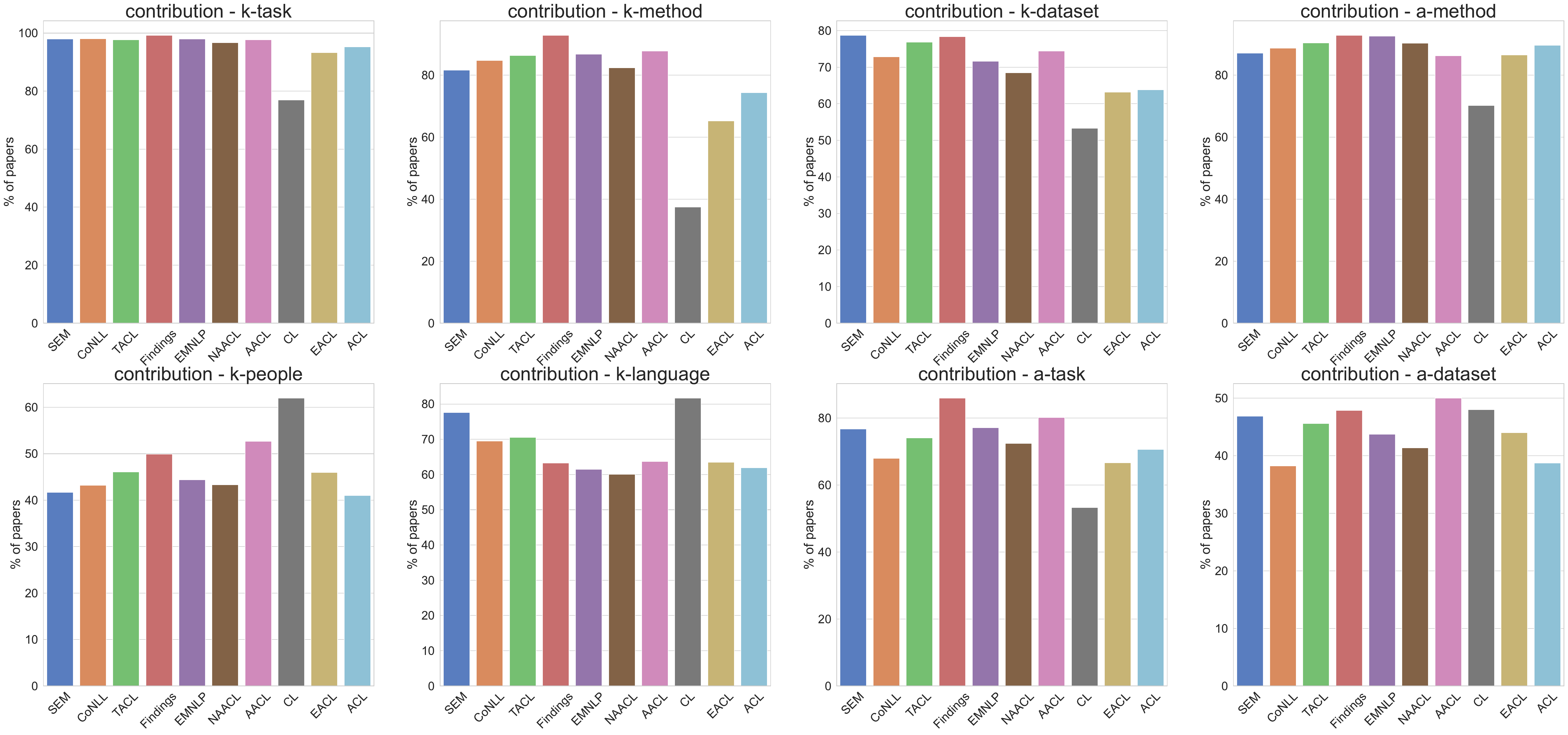}
    }
    \caption{Distribution of contribution types across research papers by conference (Abbr.: knowledge (k), artifact (a)).}
    \label{fig:conf_dist}
     %\vspace*{-3mm}
\end{figure*}

% \begin{figure}
%     \centering
%     \scalebox{0.95}{
%     \includegraphics[width=1\columnwidth, height=26.0cm]{asssets/temporal_conf_contrib_dist_v2.pdf}
%     }
%     \caption{Development in the distribution of contribution types across research papers by conference (Abbr.: knowledge (k), artifact (a)). Refer to Figure~\ref{fig:temp_conf_dist_p1}, ~\ref{fig:temp_conf_dist_p2},~\ref{fig:temp_conf_dist_p3} for larger figures.}
%     \label{fig:temp_conf_dist}
%      %\vspace*{-3mm}
% \end{figure}

\begin{table*}[!ht]
    \centering
    \begin{adjustbox}{width=2.0\columnwidth, center}
        \begin{tabular}{l l p{16.0cm}}
        \toprule
        {\bf Type} & {\bf Sub-type} & {\bf Prompt}\\
        \midrule
        \multirow{5}{*}{Knowledge} & k-task & Central to NLP research are tasks such as Machine Translation, Named Entity Recognition, Language Modeling, etc. Your task is to assess whether the provided sentence from an NLP research paper describes new knowledge about any of such existing NLP tasks, including new knowledge about their properties or characteristics. However, the sentence should not propose a new NLP task. Respond with "yes" if the sentence presents new knowledge about one or more of these NLP tasks; otherwise, respond with "no".\\
        \hline
        & k-method & NLP Models such as RNNs, LSTMs or LLMs are indispensable for NLP Research. Your task is to determine if the provided sentence from an NLP research paper describes new knowledge or analysis about such existing NLP models or methods like RNNs, LSTMs, or LLMs. However, the sentence should not propose new models or methods. Respond with "yes" if the sentence presents new knowledge about NLP models; otherwise, respond with "no."\\
        \hline
        & k-people & In NLP research, every paper plays a role in advancing the field. Your task is to assess whether the given sentence from an NLP research paper presents new knowledge about people, humankind, society or human civilization. Respond with "yes" if the sentence describes novel knowledge about people, humankind, society or human civilization; otherwise, respond with "no." Use only a yes or no format for your answers.\\
        \hline
        & k-dataset & Datasets constitute a crucial aspect of NLP and machine learning research. Examining datasets can yield valuable insights into their properties and features. Your task is to assess whether the given sentence from an NLP research paper describes new knowledge about a dataset, such as its new properties or characteristics or describes new knowledge concerning properties or characteristics of datasets in general. Respond with "yes" if the sentence presents novel knowledge about the datasets; otherwise, respond with "no." Use only a yes or no format for your answers. \\
        \hline
        & k-language & In NLP research, every paper plays a role in advancing the field. Your task is to assess whether the given sentence from an NLP research paper presents new knowledge about language, such as a new property or characteristic of language. Respond with "yes" if the sentence describes novel knowledge about language; otherwise, respond with "no."  Use only a yes or no format for your answers. \\
        \midrule 
        \midrule
        \multirow{3}{*}{Artifact} & a-task & Central to NLP research are tasks such as machine translation, named entity recognition, sentiment classification, and more. 
        Your task is to assess if the given sentence from an NLP research paper introduces, or proposes a new or novel NLP task. 
        This new task could either build upon existing NLP tasks or could be entirely novel.
        Respond with "yes" if the sentence introduces, or proposes a new or novel NLP task; otherwise, respond with "no."\\
        \hline
        & a-method & Algorithms and NLP models such as RNNs, LSTMs or LLMs are indispensable for NLP Research. 
        Your task is to assess if the provided sentence from an NLP research paper introducing, or proposing a new or novel such NLP model, algorithm, or technique. 
        This new model could have been built on top of existing models or methods or could be a completely new model.
        Respond with "yes" if the sentence introduces or proposes a new or novel NLP model; otherwise, respond with "no."\\
        \hline
        & a-dataset & Datasets constitute a crucial aspect of NLP research.
        Your task is to assess whether the given sentence from an NLP research paper introduces or discusses a new or novel NLP dataset.
        Respond with "yes" if it does; otherwise, respond with "no."  \\

        \bottomrule
        
        \end{tabular}
    \end{adjustbox}
    \caption{Prompts to identify different types of contributions from NLP Research papers using LLMs.}
    %\caption{Annotation Scheme[TODO: improve formatting]}
    \label{tab:llm_prompts}
\end{table*}

\begin{figure}
    \centering
    \scalebox{0.9}{
    \includegraphics[width=0.5\textwidth]{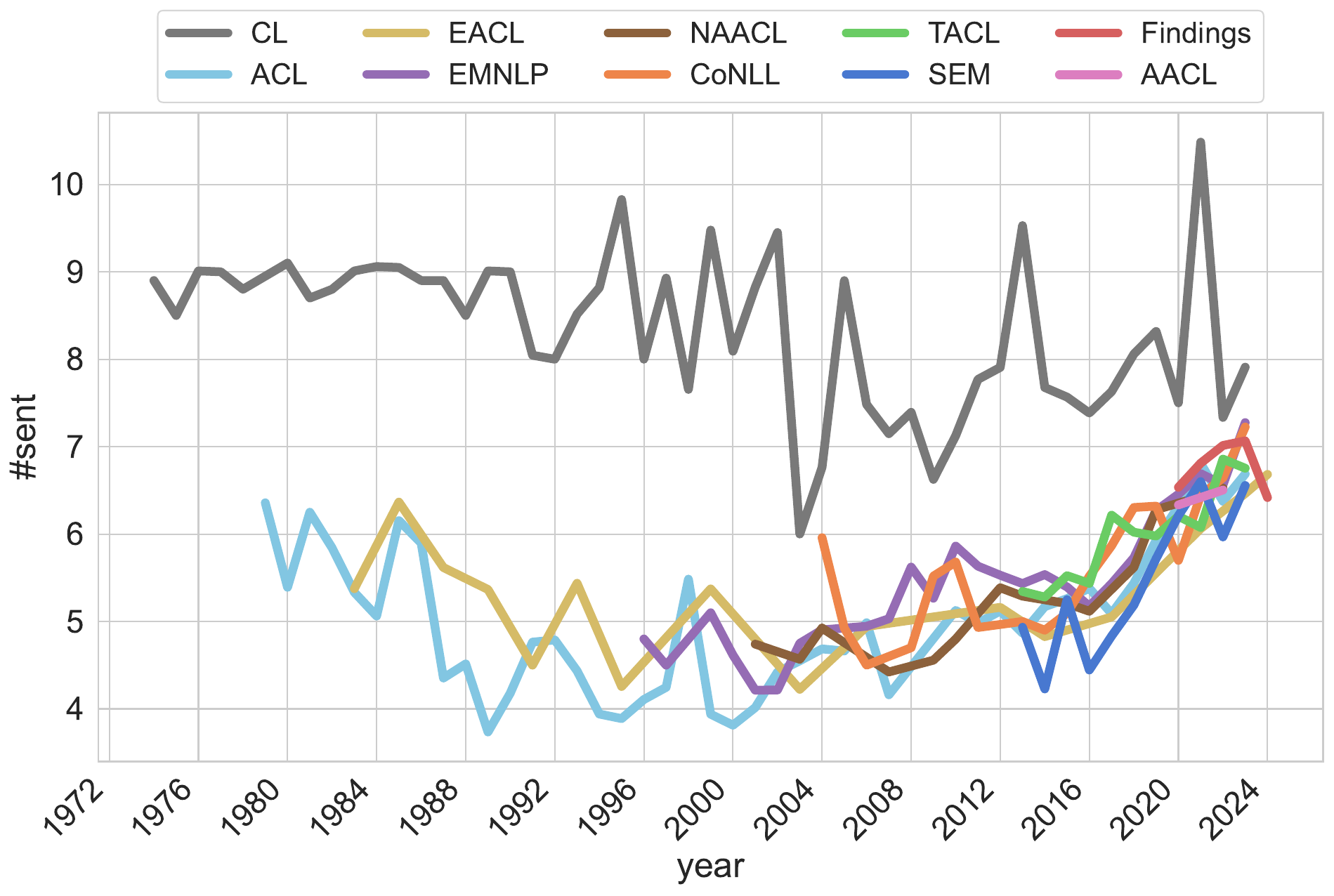}
    }
    \caption{Average abstract length in papers from different venues.}
    % \caption{Average length of abstract per paper across different venues.}
    \label{fig:conf_abs}
     %\vspace*{-3mm}
\end{figure}

\begin{figure}
     \centering
     \begin{subfigure}[b]{0.5\textwidth}
         \centering
         \scalebox{0.90}{
         \includegraphics[width=\textwidth]{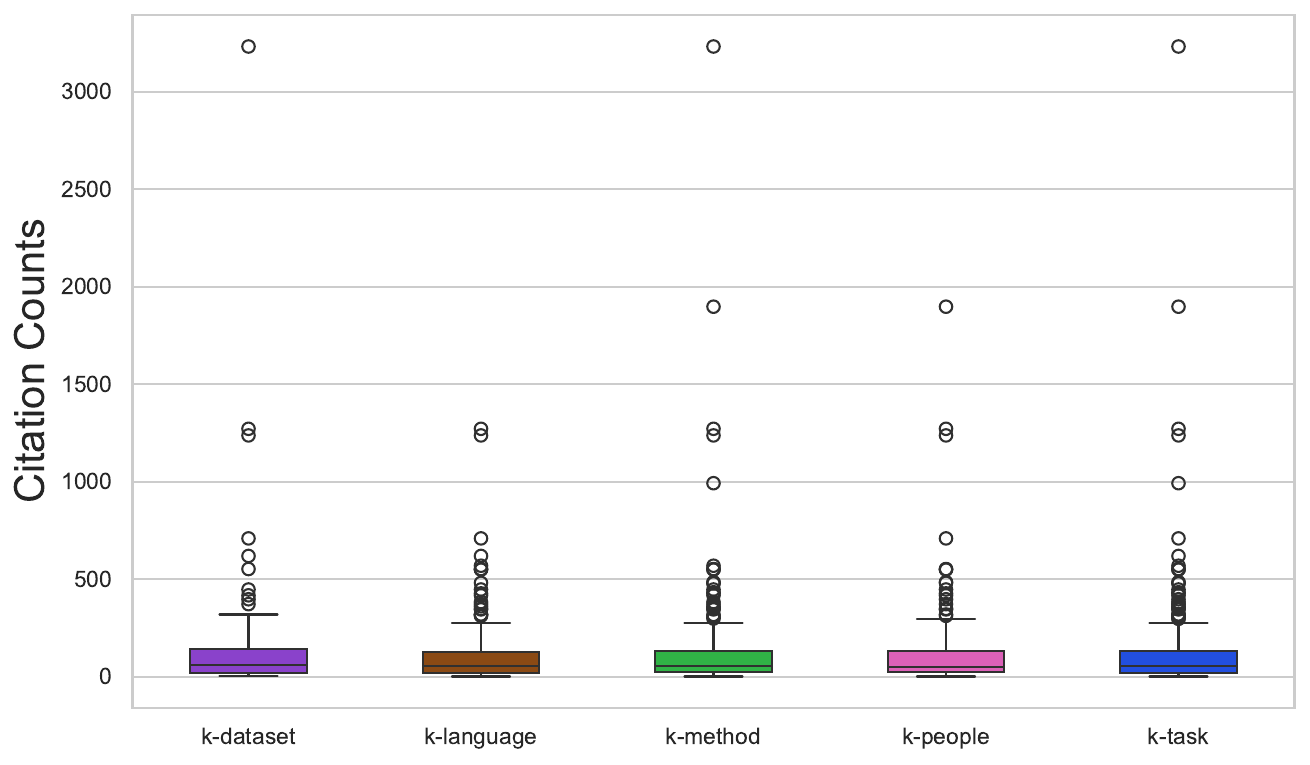}
         }
         \caption{Knowledge contributions}
         \label{subfig:k_boxplot}
     \end{subfigure}
     \hfill
     %\hspace{1.5cm}
     \begin{subfigure}[b]{0.5\textwidth}
         \centering
         \scalebox{0.90}{
         % \includegraphics[width=\textwidth]{asssets/temporal_a_cumulative_label_dist.pdf}
         % }
         \includegraphics[width=\textwidth]{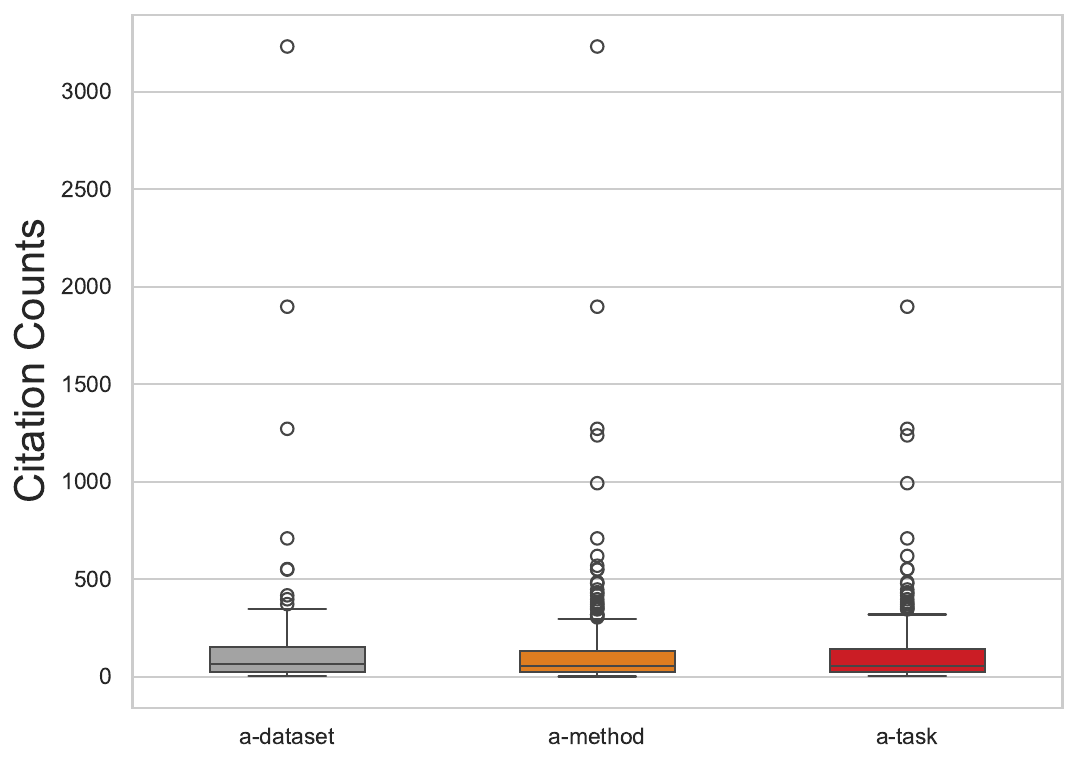}
         }
         \caption{Artifact contributions}
         \label{subfig:a_boxplot}
     \end{subfigure}
     % \hfill
     % \begin{subfigure}[b]{0.3\textwidth}
     %     \centering
     %     \includegraphics[width=\textwidth]{graph3}
     %     \caption{$y=5/x$}
     %     \label{fig:five over x}
     % \end{subfigure}
        \caption{Variability and asymmetry in citation counts for each contribution type.}
        % \caption{Distribution of Citation Counts Across Contribution Types.}
        \label{app_fig:citation_boxplot}
        %\vspace*{-5mm}
\end{figure}

% \begin{figure*}
%     \centering
%     \scalebox{1.0}{
%     % \includegraphics[width=0.5\textwidth]{asssets/conf_avg_abs.pdf}
%     \includegraphics[width=0.65\textwidth]{asssets/conf_avg_abs_v2.pdf}
%     }
%     \caption{Average length of abstract per paper across different venues.}
%     \label{fig:conf_abs}
%      %\vspace*{-3mm}
% \end{figure*}

% \begin{figure*}
%     \centering
%     \scalebox{1}{
%     \includegraphics[width=1\textwidth]{asssets/conf_contrib_dist_v2.pdf}
%     }
%     \caption{Distribution of contribution types across research papers by conference (Abbr.: knowledge (k), artifact (a)).}
%     \label{fig:conf_dist}
%      %\vspace*{-3mm}
% \end{figure*}

\begin{figure*}
    \centering
    \scalebox{1}{
    \includegraphics[width=1\textwidth]{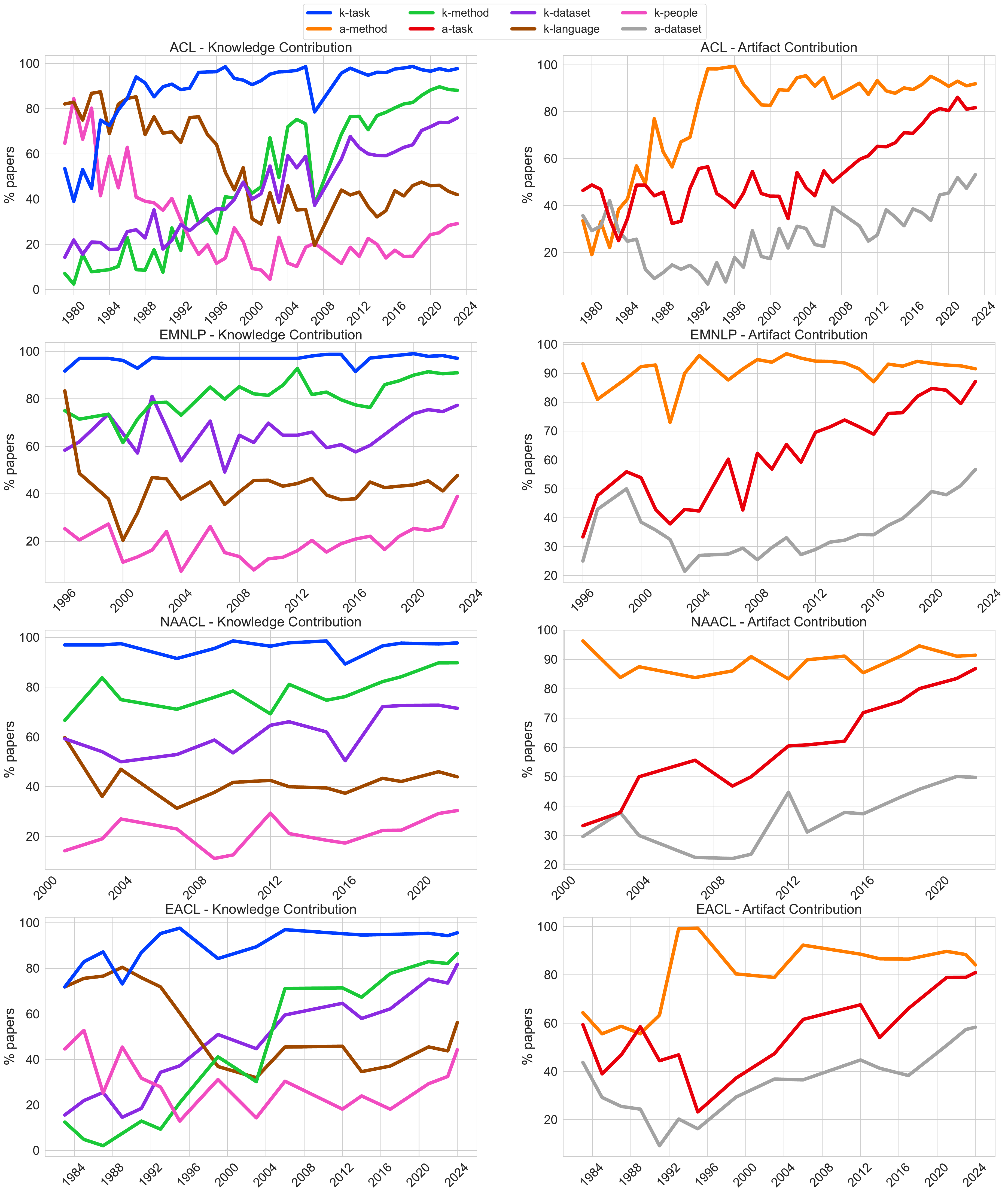}
    }
    \caption{Evolution of the four venues (ACL, EMNLP, NAACL, and EACL) based on the percentage of papers containing at least one contribution of each type (Abbr.: knowledge (k), artifact (a)).}
    \label{fig:temp_conf_dist_p1}
     %\vspace*{-3mm}
\end{figure*}

\begin{figure*}
    \centering
    \scalebox{1}{
    \includegraphics[width=1\textwidth]{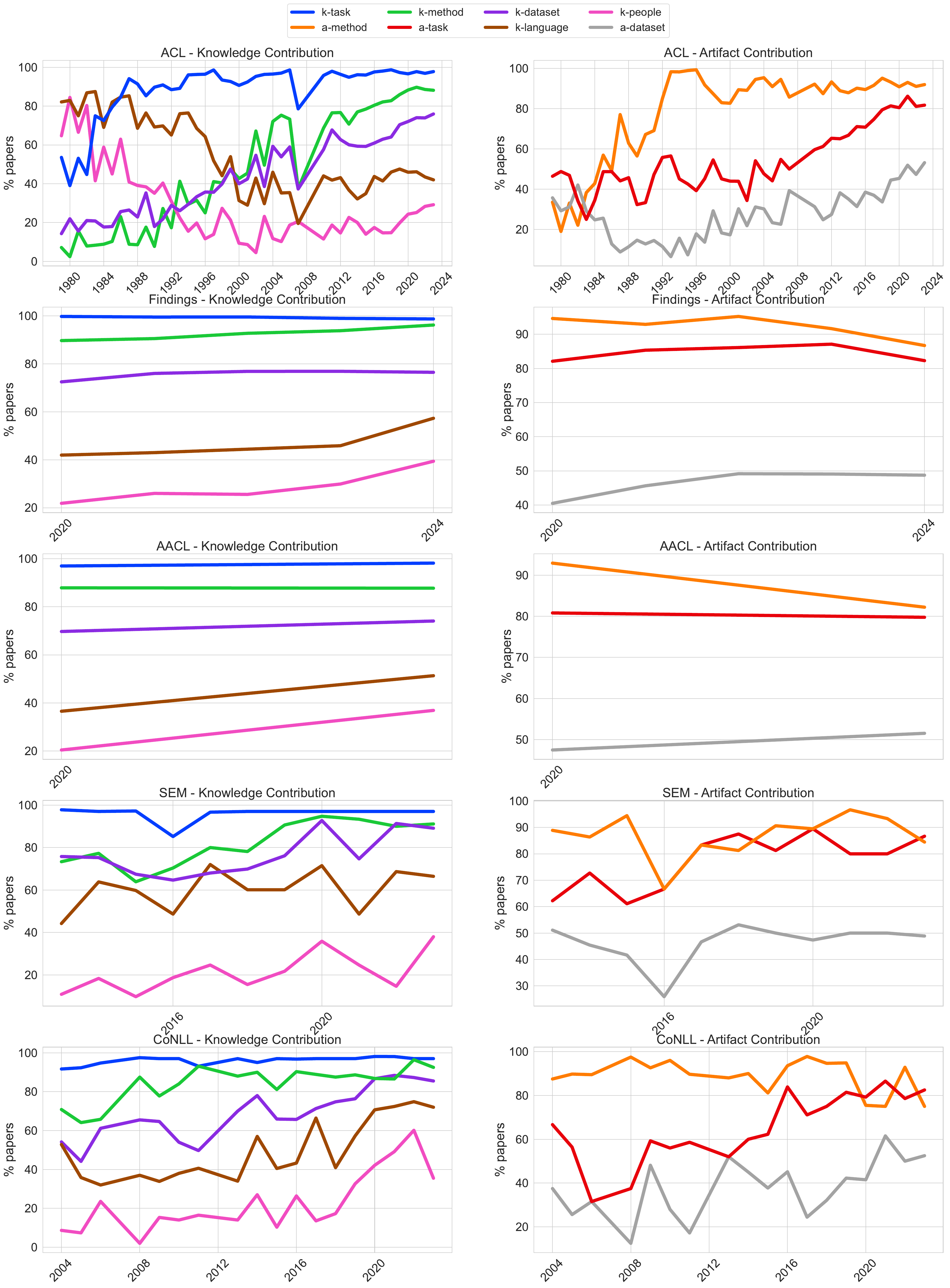}
    }
    \caption{Evolution of the five venues (ACL, Findings, AACL, *SEM, and CoNLL) based on the percentage of papers containing at least one contribution of each type (Abbr.: knowledge (k), artifact (a)).}
    \label{fig:temp_conf_dist_p2}
     %\vspace*{-3mm}
\end{figure*}

\begin{figure*}
    \centering
    \scalebox{1}{
    \includegraphics[width=1\textwidth]{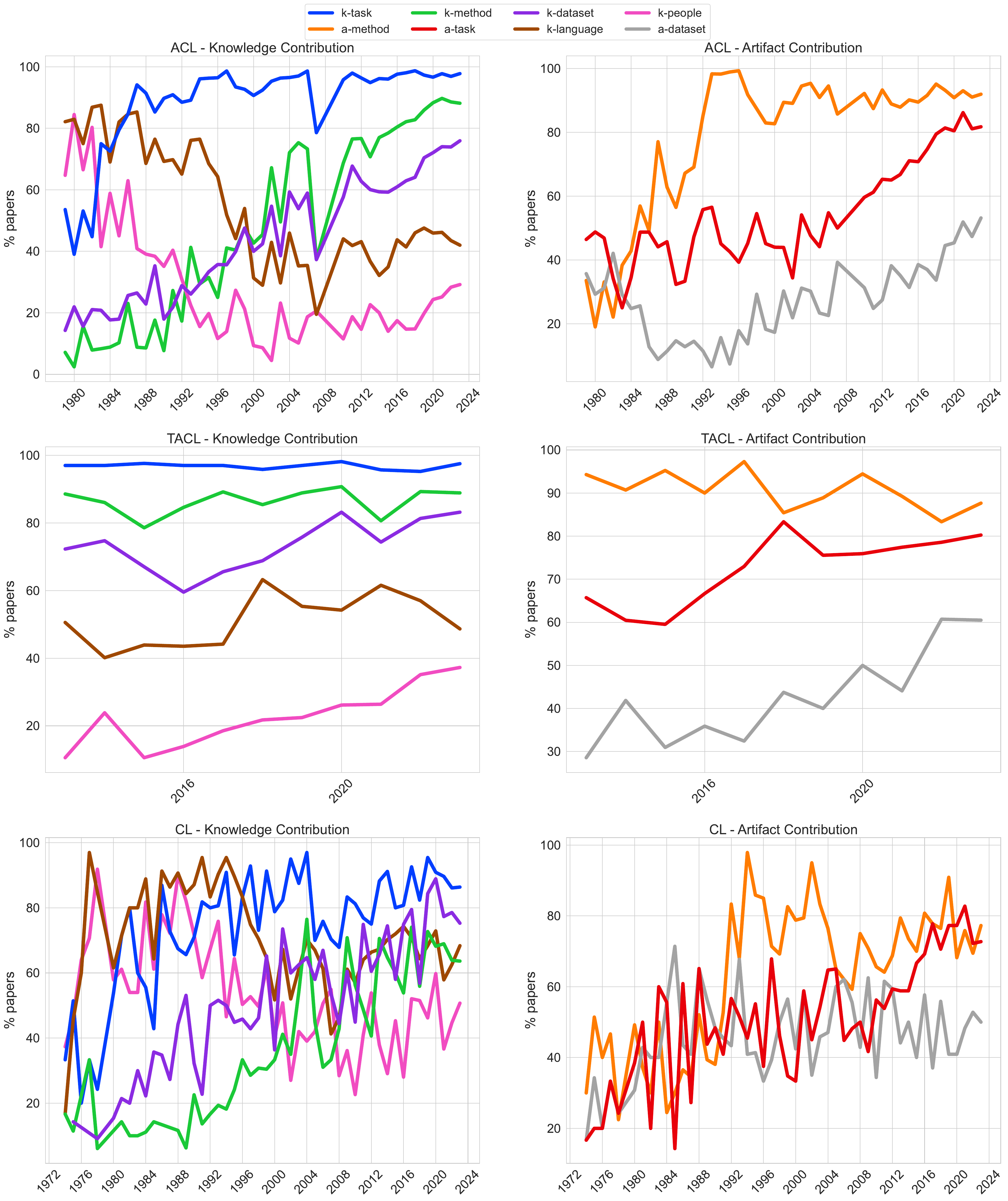}
    }
    \caption{Evolution of the three venues (ACL, TACL, and CL) based on the percentage of papers containing at least one contribution of each type (Abbr.: knowledge (k), artifact (a)).}
    \label{fig:temp_conf_dist_p3}
     %\vspace*{-3mm}
\end{figure*}